%% file: hsics.tex
\documentclass[10pt,a4paper]{article} 

\usepackage[compress]{cite}

\usepackage{booktabs}

\usepackage[utf8]{inputenc}
\usepackage{times}

\usepackage[normalem]{ulem}

\usepackage{graphicx,adjustbox,subcaption}
\usepackage{amsmath,mathtools,amsfonts,amssymb,units,bm}
\usepackage{algorithm,algorithmicx}
\usepackage{amsthm}
\usepackage[noend]{algpseudocode}
\usepackage{enumitem}
\setlist[itemize,1]{leftmargin=\dimexpr 26pt-0.5cm}

\usepackage{acronym}
\usepackage{todonotes,comment} 
\usepackage{hyperref} 

\usepackage{enumitem}
\usepackage{fullpage,afterpage}

\DeclareMathAlphabet\boldsymbolcal{OMS}{cmsy}{b}{n}

\hypersetup{
	colorlinks=true,
	linkcolor=black,      
	citecolor=black,      
	filecolor=black,      
	urlcolor=black        
}

\input{acro.tex}

\graphicspath{{./fig/}}

\title{Multispectral Compressive Imaging Strategies \\ using Fabry-P\'{e}rot Filtered Sensors}
\author{
	Kévin Degraux, Valerio Cambareri, Bert Geelen, \\Laurent Jacques, Gauthier Lafruit
	\thanks{KD, VC, and LJ are with the ISPGroup ICTEAM/ELEN, Universit\'e catholique de Louvain, Belgium, LJ is funded by the F.R.S.-FNRS, part of this work has been funded by the \emph{p\^ole Mecatech} Walloon region project ADRIC (e-mail: \{kevin.degraux, valerio.cambareri, laurent.jacques\}@uclouvain.be). GL is with LISA, Universit\'e libre de Bruxelles (e-mail: \url{gauthier.lafruit@ulb.ac.be}). BG is with imec, Belgium (e-mail: \url{bert.geelen@imec.be}). 
	Computational resources have been provided by the CISM/UCL and the C\'ECI in FWB funded by the F.R.S.-FNRS under convention 2.5020.11.}%
}

\input{macros.tex}

\begin{document} 
	\maketitle
	
	\input{introhsi.tex}

	\section{Preliminaries}
	\input{preliminaries.tex}

	\section{Multispectral Compressive Imaging by Generalized Inpainting}
	\input{inpaint.tex}

	\section{Multispectral Compressive Imaging by Out-of-Focus Random Convolution}
	\input{msrconv.tex}

	\section{Numerical Experiments}
	\input{comp_exp.tex}

	\section{Conclusion}
	\input{conclhsi.tex}

	\bibliographystyle{IEEEtran_mod}
	\bibliography{library,HSIbib,TI_DLP}
\end{document}

%% file: acro.tex
\newacro{ADMM}{Alternating Directions Method of Multipliers}
\newacro{AWGN}{Additive White Gaussian Noise}
\newacro{BPDN}{Basis Pursuit with Denoising}
\newacro{CA}{Coded Aperture}
\newacro{CASSI}{Coded Aperture Snapshot Spectral Imaging}
\newacro{CS}{Compressed Sensing}
\newacro{DCT}{Discrete Cosine Transform}
\newacro{DFT}{Discrete Fourier Transform}
\newacro{FFT}{Fast Fourier Transform}
\newacro{DOF}{depth-of-field}
\newacro{DWT}{Discrete Wavelet Transform}
\newacro{FP}{Fabry-P\'erot}
\newacro{FPA}{Focal Plane Array}
\newacro{HS}{hyperspectral}
\newacro{IHT}{Iterative Hard Thresholding}
\newacro{iid}[i.i.d.]{independent and identically distributed}
\newacro{lsc}[l.s.c.]{lower semi-continuous}
\newacro{MS}{multispectral}
\newacro{MSE}{Mean Squared Error}
\newacro{MSRC}{Multispectral Random Convolution}
\newacro{MSVI}{Multispectral Volume Inpainting}
\newacro{PFE}{Partial Fourier Ensemble}
\newacro{PMF}{Probability Mass Function}
\newacro{PSF}{Point Spread Function}
\newacro{PSNR}{Peak Signal-to-Noise Ratio}
\newacro{RIC}{Restricted Isometry Constant}
\newacro{RIP}{Restricted Isometry Property}
\newacro{RME}{Random Matrix Ensemble}
\newacro{ROI}{region of interest}
\newacro{RV}{Random Vector}
\newacro{SLM}{Spatial Light Modulator}
\newacro{SNR}{Signal-to-Noise Ratio}
\newacro{TV}{Total Variation}
\newacro{UDWT}{Undecimated Discrete Wavelet Transform}
\newacro{URA}{Uniformly Redundant Array}
\newacro{VIS}{visible light}
\newacro{FWHM}{Full Width at Half Maximum}


%% file: macros.tex
\newcommand{\NN}{\mathbb{N}}
\newcommand{\CC}{\mathbb{C}}

\newcommand{\RR}{\mathbb{R}}

\newcommand{\Cc}{\mathcal{C}}

\newcommand{\Rr}{\mathcal{R}}

\newcommand{\al}{\alpha}
\newcommand{\la}{\lambda}

\newcommand{\de}{\delta}
\newcommand{\De}{\Delta}
\renewcommand{\phi}{\varphi}

\newcommand{\Om}{\Omega}

\DeclareMathOperator{\diag}{diag}
\DeclareMathOperator{\bdiag}{bdiag}

\DeclareMathOperator{\prox}{prox}

\DeclareMathOperator{\rect}{rect}
\DeclareMathOperator{\sinc}{sinc}
\DeclareMathOperator{\vvec}{vec}

\def\eqdef{\triangleq}

\newcommand{\norm}[1]{\left\| #1 \right\|}
\newcommand{\abs}[1]{\left| #1 \right|}
\newcommand{\ens}[1]{\left\{ #1 \right\}}
\newcommand{\pa}[1]{\left( #1 \right)}

\newcommand{\enscond}[2]{ \left\{ #1 \;\middle|\; #2 \right\} }

\def\dif{{\rm d}}

\DeclareMathOperator{\argmin}{argmin}

\newcommand{\uargmin}[1]{\underset{#1}{\argmin}\;}

\newcommand{\qandq}{ \quad \text{and} \quad }

\newcommand{\ie}{\emph{i.e.}, }
\newcommand{\eg}{\emph{e.g.}, }

\renewcommand{\vec}[1]{{\boldsymbol #1}}
\newcommand{\mat}[1]{{\boldsymbol #1}}
\newcommand{\ten}[1]{{\boldsymbol {\mathcal #1}}}

\newcommand{\Id}{\mathbf{I \mkern-1mu d} }

\newcommand{\zero}{\vec{0}}

\newcommand{\vx}{\vec{x}}
\newcommand{\vy}{\vec{y}}

\newcommand{\vv}{\vec{v}}
\newcommand{\vw}{\vec{w}}
\newcommand{\vz}{\vec{z}}

\newcommand{\val}{\vec{\al}}

\newcommand{\mPhi}{\mat{\Phi}}

\newcommand{\mA}{{\bf A}}

\newcommand{\mF}{{\bf F}}

\newcommand{\Rn}{\mathbf{R}_n}
\newcommand{\Rm}{\mathbf{R}_m}
\newcommand{\Rnc}{\mathbf{R}_n^c}

\newcommand{\Up}{\mathbf{ U \mkern-2mu p  }}

\newcommand{\xz}{\vx_{0}}

\newcommand{\xh}{\hat{\vx}}

\theoremstyle{remark}

\newcommand{\eq}[1]{\begin{equation*}
#1
\end{equation*}
}

\newcommand{\eql}[1]{\begin{equation}
#1
\end{equation}
}

\newcommand{\eqn}[1]{\begin{align*}
#1
\end{align*}
}

\def\dif{{\rm d}}

\def\sinc{{\rm sinc}}
\def\rect{{\rm rect}}



%% file: introhsi.tex
	\begin{abstract}
		This paper introduces two acquisition device architectures for multispectral compressive imaging.
Unlike most existing methods, the proposed computational imaging techniques do not include any dispersive element, as they use a dedicated sensor which integrates narrowband Fabry-Pérot spectral filters at the pixel level.
%
The first scheme leverages joint inpainting and super-resolution to fill in those voxels that are missing due to the device's limited pixel count.
The second scheme, in link with compressed sensing, introduces spatial random convolutions, but is more complex and may be affected by diffraction.
%
In both cases we solve the associated inverse problems by using the same signal prior. Specifically, we propose a redundant analysis signal prior in a convex formulation. 
Through numerical simulations, we explore different realistic setups. Our objective is also to highlight some practical guidelines and discuss their complexity trade-offs to integrate these schemes into actual computational imaging systems. 
Our conclusion is that the second technique performs best at high compression levels, in a properly sized and calibrated setup. Otherwise, the first, simpler technique should be favored.
	\end{abstract}
	
		Multispectral imaging, compressed sensing, spectral filters, Fabry-P\'erot, random convolution, generalized inpainting.

	\section{Introduction}
	\label{sec:hsics_intro}
	Multispectral (\acsu{MS}) imaging consists in capturing the light intensity, $X_0(u,v,\lambda)$,
of an object or scene as it varies along its 2-D spatial coordinates $(u,v)$ and over different wavelengths $\lambda$, \ie the light spectrum as measured into a few intervals or {\em bands}. 
	This information is sampled in a 3-D data volume, which allows for accurate classification or segmentation of constituents in an object or scene from their spectral profile. Hence, \ac{MS} imaging finds diverse applications in remote sensing \cite{Shippert2004}, optical sorting \cite{Tatzer2005}, astronomy \cite{Hege2004}, food science \cite{Gowen2007}, medical imaging \cite{Lu2014} and precision agriculture \cite{Whiting2006}.

	A classic approach is to spatially or spectrally multiplex the \ac{MS} cube over a 2-D \ac{FPA}. This is done by \emph{scanning} the cube, so that specific \emph{slices} are sequentially acquired by the sensor in several snapshots (for a review see, \eg \cite{SellarBoreman2005}). Such systems require either tunable spectral filters or dispersive elements with mechanical parts to scan the object or scene. These approaches entail trade-offs between complexity and cost, spectral and spatial resolution, and acquisition time. 

	Recently, single-snapshot \ac{MS} imagers were developed to rapidly acquire a \ac{MS} cube, thus avoiding motion artifacts and enabling video acquisition rate \cite{HagenKudenov2013}. 
	Among such imagers, we focus on those using \ac{FP} filtered sensors~\cite{lambrechts2014cmos,geelen2014compact}, \ie standard CMOS imaging sensors on top of which an array of spectral filters is deposited. 
	This technique generalizes RGB filter arrays \cite{Bayer1976} to filter banks using an arbitrary number of narrowband profiles \cite{lambrechts2014cmos}, \eg a few tens.  
	Thus the array imposes a reduction in spatial resolution as the sensor's pixels are partitioned between bands.
	
	This paper investigates \ac{MS} imaging strategies based on \ac{CS} (see, \eg \cite{Donoho2006,CandesWakin2008}), an established signal processing paradigm that has inspired several computational imaging frameworks~\cite{WillettMarciaNichols2011,ArceBradyCarinEtAl2014,GehmJohnBradyEtAl2007,GehmBrady2015}.
	After acquisition by a compressive device, the measurements are fed into a {\em recovery algorithm} along with the \emph{sensing operator} and \emph{signal prior}. 
	Under broad theoretical conditions \cite{Candes2008,Rauhut2010a}, this method recovers a high-resolution approximation of the target scene, even if the sensing was performed below the scene's Nyquist rate. 
	The complexity of the sensing operation (\eg resolution, time) is therefore balanced to the complexity of the signal with respect to a given prior.

	\subsection{Main Contributions}

	Our work contributes to advancing the field of \ac{MS} compressive imaging in the following senses:

	\begin{enumerate}[label=(\roman*)]
	\item We propose two \ac{MS} snapshot imaging strategies: \ac{MSVI} and \ac{MSRC}. Both maintain a relatively low system-level complexity \emph{without any dispersive element}. Using \ac{CS} principles, they are designed with a low-pixel-count \ac{FP} sensor. 
	
	\item \ac{MSVI} leverages a \emph{generalized inpainting} procedure, as discussed in~\cite{Degraux2015a}, to provide a simple integration of the \ac{FP} sensor in a computational imaging scheme.  
This architecture performs a spatio-spectral subsampling of the \ac{MS} cube and relies on their redundancy to obtain a high-resolution recovery. It is fairly simple and works best at lower compression levels.
	
	\item \ac{MSRC} leverages \emph{random convolution}, as discussed in~\cite{Degraux:iTWIST14}, to provide spatial-domain \ac{CS} by means of an out-of-focus random \ac{CA}, \ie an array of square apertures randomly placed on an opaque screen. It preserves the spectral resolution, fixed by the low number of narrowband \ac{FP} filters (\eg $16$) on the \ac{FPA}. In an ideally sized, low-noise setup, this more complex architecture clearly improves the recovered quality, especially at higher compression levels. However, it entails some optical design challenges, as discussed in Section~\ref{sec:msrc-nonidealities}.
	
		\item Our analysis is paired with a discussion on the analysis-sparse signal prior, the associated convex optimization formulation and fine-tuned \acs*{ADMM} algorithm \cite{Boyd2010, Almeida2013} for the large-scale recovery of \ac{MS} cubes.
	
	\item Both architectures are numerically compared in terms of achievable recovery performances. We also discuss their complexity trade-offs and design guidelines, by identifying unavoidable adverse optical effects, to integrate these schemes into realistic imaging systems.

\end{enumerate}

	Our findings and numerical results corroborate how a conspicuous reduction in the number of measurements w.r.t.  the Nyquist-rate representation of $X_0(u,v,\la)$ is made possible by both architectures while preserving high \ac{PSNR}. Table~\ref{table:comparison} summarizes some pros and cons of each strategy, which are detailed and clarified throughout the paper. 
	
The paper is organized as follows. The rest of Section~\ref{sec:hsics_intro} places our contribution in perspective with respect to related work in the literature and introduces useful notations. Section~\ref{sec:preliminaries} presents the \ac{FP} filters technology, the proposed \ac{MS} analysis sparsity prior and the inverse problem formulation, common to both strategies, as well as the associated reconstruction algorithm. Section~\ref{sec:msvi} details the specifics of the \ac{MSVI} strategy, \ie its image formation model and sensing matrix. Section~\ref{sec:msrc} similarly provides the details of the \ac{MSRC} strategy and discusses some associated non-idealities and practical considerations. Section~\ref{sec:numerics} presents numerical reconstruction results. We demonstrate the \ac{MSVI} performances with experimental data and compare \ac{MSVI} and \ac{MSRC}, using simulated acquisition. The final section gives a brief conclusion.

\begin{table}[t]
\centering

\begin{tabular}{lll}
  \toprule
  & \textbf{MSVI}  & \textbf{MSRC} \\ \midrule
  Optical system complexity    & Very simple & Simple (simpler than \cite{GehmJohnBradyEtAl2007})\\
  Optical calibration                 & Simple & Complex\\
  Robustness to noise &  Good & Good  \\
  Robustness to miscalibration &  Acceptable & Low (see, \eg~\cite{Cambareri2016,Ling2015}) \\
  PSNR at 1:16 compression     & 33dB & 37dB (with or without PSF)   \\
  PSNR at 1:2 compression      & 48dB & 50dB (41dB with PSF) \\

  Acquisition speed      &  Fast & Fast \\
  Initialization  quality  \eqref{eq:tikhonov_x0}        &  Acceptable & Low\\
  \bottomrule
  \end{tabular}
  \caption{\label{table:comparison} Comparison of the proposed imaging strategies. \acf{PSNR} based on Fig.~\ref{fig:quantitative_result}. See \ac{MSVI} initialization on Fig.~\ref{fig:patches_results}.}
\end{table}

	\subsection{Related Work}
	\label{sec:hsics_related_work}
	\subsubsection{Compressive Spectral Imaging} 
	The use of \ac{CS} for \ac{MS} imaging schemes dates back to \cite{GehmJohnBradyEtAl2007,SunKelly2009}. The most popular application of \ac{CS} to spectral imaging is the \ac{CASSI} framework \cite{GehmJohnBradyEtAl2007, WagadarikarPitsianisSunEtAl2009,WagadarikarJohnWillettEtAl2008,WuMirzaArceEtAl2011,ArceBradyCarinEtAl2014}, with its many variants summarized hereafter. Single-disperser \ac{CASSI} uses a random \ac{CA} to partially occlude the focused scene. A refractive prism or grating then shears the spatio-spectral information, and the processed light is recorded by a standard imaging sensor. The system introduced by \cite{WagadarikarJohnWillettEtAl2008} shows high spectral accuracy after image recovery, at the expense of lower spatial accuracy. Double-disperser \ac{CASSI} \cite{GehmJohnBradyEtAl2007} achieves opposite performances in terms of spatial {\em versus} spectral accuracy, but requires non-trivial calibration and geometric alignment of its optical components.  
	A close line of work in \cite{CorreaArguelloArce2014,CorreaArguelloArce2015} proposes a snapshot spectral imaging architecture. It is based on \ac{CASSI} but features wide-band spectral filters, which provide random spatio-spectral subsampling after the shearing element. Non-snapshot spectral imaging architectures based on \ac{CS} were also recently proposed~\cite{SunKelly2009,AugustVachmanRivensonEtAl2013,Fowler2014}.
	
	\ac{CASSI} and its variants target a relatively large number of bands and are intrinsically capable of achieving high spectral resolution thanks to dispersive optics. However, when the spectrum is well represented by fewer bands, spectral mixing is less effective, for \ac{CS} purposes, than spatial mixing,
	especially for \ac{FP}-filtered sensors with only a few tens of narrowband filters (\eg \cite{GeelenBlanchGonzalezEtAl2015}) whose high selectivity excludes spectral super-resolution. 
	In this work, we focus on those \ac{FP} filtered schemes that target a few bands without using any dispersive element.
		
	\subsubsection{Compressive Imaging by Random Convolution}
		Since its introduction, \ac{CS} has been envisioned to provide image acquisition at reduced sensor resolution \cite{duarte:2008} or shorter acquisition times \cite{Lustig2007} (see also the tutorial in \cite{willett2011compressed} and references therein). In particular, the second strategy proposed in this paper is related to \ac{CS} by {random convolution}: The sensing operation acts as a spatial-domain convolution with a random filter, \eg a random \ac{CA} as in \ac{CA} imaging \cite{dicke1968scatter,fenimore1978coded}. 
	More recently, the subsampled random convolution operation was shown, in \cite{Romberg2009,Rauhut2010a, Rauhut2012}, to comply with theoretical results of \ac{CS}. This operation was also featured in recent imaging architectures \cite{Jacques2009, Marcia2009, Bjorklund2013}. Convolution-based schemes are appealing because they allow for a fast sensing operation. Indeed, the compressive measurements can be formed in one frame of a full imaging sensor, as opposed to {\em single-pixel camera} designs \cite{duarte:2008, SunKelly2009, chan2008single}, where the compressive measurements are  multiplexed in time. Moreover, \ac{FFT} implementations of the convolutions drastically reduce the computational cost of reconstruction compared to unstructured sensing operations.
	However, the snapshot capability and numerical efficiency of random convolution architectures are paid by a higher correlation between adjacent compressive measurements, because of their spatial adjacency and considering the optical-level non-idealities such as the \ac{PSF} of the optical elements.
	In this work we propose a \ac{MS} extension of the low-complexity snapshot imaging architecture introduced by Bj\"orklund and Magli \cite{Bjorklund2013}. In particular, their architecture uses a \ac{CA} placed out-of-focus  to provide random convolution.

\subsection{Notations}
\label{sec:notations}
Vectors are noted $\vx \in \RR^n$ (bold), matrices, $\mat{X} \in \RR^{n_1 \times n_2}$ (upper-case bold), 3-D arrays, $\ten{X} \in \RR^{n_1 \times n_2 \times n_3}$ (caligraphic bold), $\vvec(\mat{X})$ is the vectorization of $\mat{X}$ in row-major ordering,
$\Id$ is the identity matrix and $\mPhi^*$ is the adjoint of $\mPhi$. We note $[n] \eqdef \{1,\dots,n\}$, $\norm{\vx}_p \eqdef \pa{\sum_{i=1}^{n} \abs{\vx}^p }^{1/p}$ is the $\ell_p$-norm of $\vx$  and $\norm{\mPhi}$ is the operator $\ell_2$-norm. The proximal operator associated to $f$ is defined by $\prox_{f}(\vv) \eqdef \argmin_{\vx} f(\vx) +\frac{1}{2} \norm{\vx- \vv}_2^2$. The \emph{full} (2-D discrete) convolution between $\mat{X}$ and $\mat{Y}$ is $\mat{X} * \mat{Y}$ and its \emph{valid} part (\textsc{Matlab} and NumPy terminology, \ie fully overlapping inputs without zero-padded edges) is $\mat{X}\ \bar{*}\ \mat{Y}$. The indicator $\iota_\Cc(\vx)\eqdef0$ if $\vx$ is inside the set $\Cc$ and $+\infty$ otherwise.
We note 
$\diag(\vx)$, the diagonal matrix with diagonal $\vx$;
$\bdiag(\mat{A},\mat{B}, \dots)$, the block-diagonal matrix with blocks  $\mat{A},\mat{B} , \dots$ arranged without overlap and $\bdiag_n(\mat{A}) \eqdef \bdiag(\mat{A},\mat{A}, \dots)$, repeating $n$ times $\mat{A}$.

%% file: preliminaries.tex
\label{sec:preliminaries}

In this section, we introduce the \ac{FP} filtered sensors used in both strategies. We then propose an \ac{MS} image analysis sparsity prior. This prior is used to regularize an inverse problem through a convex optimization program, again common to both strategies. This section then describes the convex formulation and the associated recovery algorithm used in Section~\ref{sec:numerics}. 

\subsection{Fabry-P\'erot Filtered Sensors}
\label{sec:fp_filters}

		\begin{figure}[t]
		\centering
		\begin{minipage}{.39\linewidth}
		\null\hfill
		\subcaptionbox{\label{fig:filter_bank}}{\includegraphics[width=\textwidth]{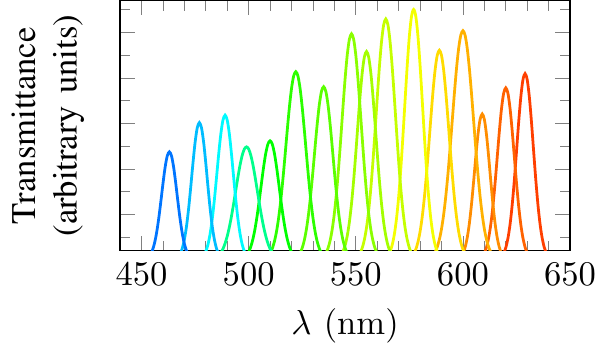}}
		\hfill\null
		\end{minipage}
		\begin{minipage}{.39\linewidth}
		\null\hfill
		\subcaptionbox{\label{fig:sensor_layouts_mosaic}}{\vspace{-0.2cm}\includegraphics[width=0.4\textwidth]{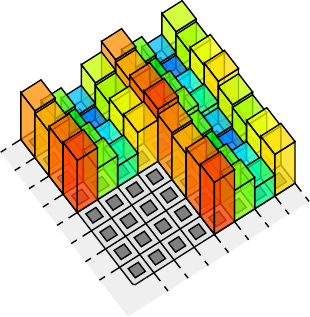}}
		\hfill\null
		\\[-2em]
		\null
		\hfill
		\subcaptionbox{\label{fig:sensor_layouts_random}}{\vspace{-0.2cm}\includegraphics[width=0.4\textwidth]{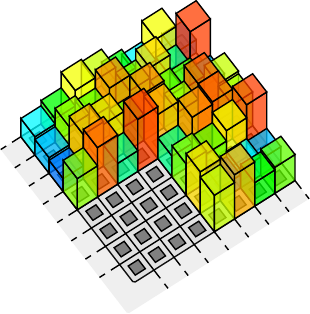}}
		\hfill
		\subcaptionbox{\label{fig:sensor_layouts_tile}}{\vspace{-0.2cm}\includegraphics[width=0.4\textwidth]{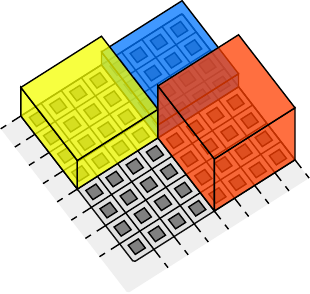}}
		\hfill\null
		\end{minipage}
		\caption{\ac{FP} filtered sensors: (a) transmittance profiles (arbitrary units) of a $16$-band \ac{FP} filter bank in the \ac{VIS} range; depiction of (b) mosaic, (c) random, and (d) tiled filters as deposited on a CMOS sensor array (gray).}
		\label{fig:filter_profiles}
		\end{figure}
		The class of imaging sensors at the core of this work are comprised of a standard CMOS sensor designed to operate in the \acf{VIS} range of wavelengths (\ie \unit[400--700]{nm}), on which a layer of \acl{FP} interferometers \cite{Fabry1901} is deposited. 
		The latter, whose physics is well described in \cite{Hernandez1986}, act on the spectrum of incoming light as band-pass filters whose center wavelength and width are designed to yield narrowband profiles (about $\unit[10]{nm}$).

		Once the filter profiles are designed, the \ac{FP} filters can be manufactured to cover the area of a single pixel, either in a {\em mosaic} layout \cite{geelen2014compact}, where a group of different filters is repeated in a $4\times 4$ or $5\times 5$ mosaic pattern (Fig.~\ref{fig:sensor_layouts_mosaic}), or by partitioning the sensor in a \emph{tiled} layout, where the sensor is partitioned in large areas with the spectral filter for a specific wavelength deposited on top of them \cite{geelen2013snapshot} (Fig.~\ref{fig:sensor_layouts_tile}). While it is possible to envision architectures that use tiled layouts \cite{Degraux:iTWIST14,geelen2014compact} we here focus on mosaic designs, as they will allow a reduction of the correlation between measurements taken on adjacent sensor pixels. 
		Such a sensor, \cite{geelen2014compact}, was designed and prototyped at imec 
		and will be referred to as \hypertarget{imec_sensor}{imec's sensor} in the following.
		The use of an external spectral cutoff filter, removing anything outside the \ac{VIS} range, allows one to obtain a filter bank such as the one depicted in Fig.~\ref{fig:filter_bank}. 
		These profiles were generated for illustration purposes based on calibration measurements taken at imec. 
		The raw data was post-processed to only keep the main lobe of each filter response. In particular, smaller secondary modes, which can appear at harmonic wavelengths (see \cite{Hernandez1986} and \cite{geelen2014compact} for another example), were removed for clarity. In this work, we consider an idealized situation, ignoring secondary modes. Furthermore, in a real situation, we must compensate for the attenuation coefficients, either before, during or after reconstruction.
		
		Hereafter, for the sake of simplicity, we approximate the spectral responses by a Dirac delta, $\delta(\lambda-\lambda_\ell)$, at each filter's centre wavelength $\lambda_\ell$, with equal gain. 
		We consider a sensor featuring a $16$-band filter bank with uniformly spaced center wavelengths between $\unit[470-620]{nm}$, either placed in a $4\times4$ mosaic pattern (Fig.~\ref{fig:sensor_layouts_mosaic}), or with a randomly-assigned arrangement called random pattern (Fig.~\ref{fig:sensor_layouts_random}). The latter has not been manufactured in practice but should not pose any major difficulty compared to the mosaic pattern. In simulations (Section~\ref{sec:msvi-simu}), the random pattern is generated by permuting the assigned locations of the filters over the entire \ac{FPA}.

	\subsection{Forward model and analysis sparsity prior}
	 Let $\ten{X}_{\mkern-3mu 0} \in \mathbb{R}^{n_u \times n_v \times n_\lambda}$ represent a discretized \ac{MS} cube in its 2-D spatial and 1-D spectral domains, equivalently represented by its vectorization $\xz \eqdef \text{vec}{(\ten{X}_{\mkern-3mu 0})} \in \mathbb{R}^n, \, n = n_u n_v n_\lambda$. 
	Both studied architectures entail a noisy linear acquisition process, summarized by the following generic \emph{forward model},
	 \begin{equation}
	 \label{eq:forward_model_hsi}
	 \vy = \mPhi \xz + \vw.
	 \end{equation}
	In this model, the linear sensing operator is represented
	 in matrix form by $\mPhi \in \RR^{m\times n}$ where $m \eqdef m_u m_v$.  
	 It yields a set of {\em compressive measurements} that are captured by the sensor array, $\mat{Y} \in \RR^{m_u\times m_v}$ or in vectorized form $\vy \eqdef \mathrm{vec}(\mat{Y}) \in \RR^m$.  The noise vector $\vw \in \RR^m$ is bounded in $\ell_2$-norm by~$\tau$.

		As any computational imaging system based on regularized inverse problems, our schemes must leverage a prior model for the signal being acquired.
		We here choose to use an \emph{analysis} sparsity prior (see, \eg \cite{Elad2006,Candes2011c,Nam2013}).
		Specifically, we separately apply linear transforms to the spatial and spectral domains, denoted by $\mA_{uv}$ and  $\mA_{\lambda}$. This amounts to constructing a separable transform by the Kronecker product $\mA \eqdef \mA_{uv} \otimes \mA_\lambda$.  
		For the spatial domain transform, $\mA_{uv}$, we chose a 2-D Daubechies-4 \ac{UDWT} which 
		forms a  shift-invariant, separable, and overcomplete wavelet transform~\cite{Mallat2009,Starck2007}. 
		The approximation level (scaling coefficients) is inherently not sparse as it contains the low-pass approximation of the image. 
		We found, however, that the slowly varying spatial information helps in leveraging the redundancy between bands. We thus use a 2-D \ac{DCT}, which concentrates the low-pass information in a few coefficients, making it consistent with our sparsity prior. The wavelet filters are chosen with length 8 and in 3 levels, resulting in an analysis transform $\mA_{uv} \in \mathbb{R}^{10 n_un_v \times n_un_v}$ (3 levels $\times$ 3 directions $+$ 1 approximation level).
	The \ac{DCT} is chosen for the 1-D spectral domain transform, $\mA_\la \in \RR^{n_\la \times n_\la}$,  
	given that we focus on \ac{MS} cubes from natural scenes with smooth spectral profiles.

\subsection{Recovery Method}
\label{sec:hsics-alg}
The recovery method consists in inverting~\eqref{eq:forward_model_hsi} to find an estimate $\hat{\vx}$ of the MS cube, using the analysis-sparsity prior. We use the $\ell_1$-analysis formulation from  \cite{Candes2011c}, with an additional range constraint $ \Rr \eqdef [x_{\mathrm{min}},x_{\mathrm{max}}]^n $, which reads
\begin{equation}
\label{eq:l1analysis}
\hat{\vx} \eqdef \uargmin{ \vx \in \Rr } { \norm{ \mA \vx }_1 \quad \mbox{s.t.} \quad \norm{\vy - \mPhi  \vx }_2 \leq \tau}.
\end{equation}
A good noise estimate can be used for setting the parameter~$\tau$. We solve the non-smooth convex optimization program \eqref{eq:l1analysis} using the \ac{ADMM} introduced in \cite{Glowinski1975,Gabay1976}. 
Specifically, we use the version from \cite[Algorithm~2]{Almeida2013}, recasted to solve problems of the form
\begin{equation}
\label{eq:admm2_objective}
\min_{ \vec{z} } \textstyle \sum_{j=1}^J g_j \pa{\mat{H}_j \vec{z}} ,
\end{equation}
where $g_j$ are convex lower semicontinuous functions and $\mat{H}_j$ are linear operators such that
$({\mat{H}_1}^*, \dots, {\mat{H}_J}^*)^* $ has full column rank. 
A practical implementation requires efficient computation of  the \emph{proximal operators}  \cite{Combettes2009,Parikh2013} associated to the functions~$g_j$, as well as the matrix-vector products $\mat{H}_j \vec{z}$ and ${\mat{H}_j}^* \vec{w}_j$ for arbitrary $\vec{z}$ and $\vec{w}_j$.  A crucial step of the algorithm is the matrix inversion, $ \pa{\sum_{j=1}^J \mu_j {\mat{H}_j}^*\mat{H}_j}^{-1} \vec{z}$, for some $\mu_j >0$. Any property of the matrices, $\mat{H}_j$, that can simplify that step should be exploited. In particular, the tight frame property, \ie ${\mat{H}_j}^*\mat{H}_j = \Id$; Fourier diagonalization, \ie $\mat{H}_j = \mF^* \mat{\Sigma} \mF$, where $\mF$ is the \ac{DFT} and $\mat{\Sigma}$ is diagonal; or the sparsity and separability of ${\mat{H}_j}$, are used in the following.

For their blind deconvolution problem, \cite{Almeida2013} proposes to handle the boundary conditions by adding and then masking the missing rows of the block-circulant sensing operator. 
This stabilizes the estimation, while recovering the block-circulant structure of the convolution operator, allowing Fourier diagonalization. 
Building over these ideas, as detailed for both architectures in Sections~\ref{sec:msvi-matrix} and~\ref{sec:msrc-matrix}, we can arbitrarily add rows and columns to $\mPhi$ in order to exploit one of the properties cited above. We define an \emph{extended} sensing matrix $\bar{\mPhi} \in \RR^{\bar{m} \times \bar{n}}$ (with $\bar{m}\geq m$ and $\bar{n} \geq n$) and \emph{restriction operators} $\Rm \in \{0,1\}^{m\times \bar{m}}$ and $\Rn \in \{0,1\}^{n\times \bar{n}}$, \ie that restrict input vectors (of length $\bar{m}$ and $\bar{n}$) to some arbitrarily chosen index sets (of length $m$ and $n$), such that
\begin{equation}
\label{eq:extended_Phi}
\mPhi  =\Rm \bar{\mPhi} \Rn^{\,*}.
\end{equation}
Note that the adjoint, $\Rn^{\,*}$, of the restriction, $\Rn$, is the corresponding zero-padding operator.
In addition to that factorization of $\mPhi$, it happens that the analysis transform $\mA$ introduced above is actually a scaled tight frame, \ie there exists a diagonal weighting matrix $\mat{\Om}$ such that $\mA \eqdef \mat{\Om}\tilde{\mA}$ and $\tilde{\mA}^*\tilde{\mA} = \Id$.
In order to make use of the tight frame property of $\mA$ and the factorization \eqref{eq:extended_Phi} of~$\mPhi$, we define
$
\bar{\mat{\Om}} \eqdef \bdiag (\mat{\Om} ,  \mat{0}_{\bar{n} - n})
$ 
and
$
\bar{\mA} \eqdef \Big( \,^{\tilde{\mA}\Rn}_{ \Rnc}   \Big),
$
where $\Rnc \in \{0,1\}^{ (\bar{n} - n) \times \bar{n} } $ is the complementary restriction of $\Rn$, such that $\Rn^{\,*}\Rn + {\Rnc}^* \Rnc = \Id $, and $ \mat{0}_{\bar{n} - n} \in \RR^{(\bar{n} - n) \times (\bar{n} - n)} $ is the zero matrix. The tight frame property, $\bar{\mA}^*\bar{\mA} = \Id$, is thus preserved.
Let $ \bar{\vx} \eqdef \Rn^{\,*} \vx$, \ie a zero-padded version of $\vx$, and let $\bar{\val} \eqdef \bar{\mA} \bar{\vx}$ and $\bar{\vz} \eqdef \bar{\mPhi} \bar{\vx}$. 
Note that $\norm{\bar{\mat{\Om}}\bar{\mA} \bar{\vx} }_1 = \norm{\mA \vx }_1$ and $\norm{\vy - \Rm \bar{\mPhi} \bar{\vx} }_2 = \norm{\vy - \mPhi \vx }_2$. By imposing $\Rnc \bar{\vx} = \zero$, we get the equivalent problem to \eqref{eq:l1analysis},
\begin{align}
\label{eq:l1analysis_ext}
\hat{\vx} &= \Rn \uargmin{ \bar{\vx}  \in \RR^{\bar{n}}} \norm{ \bar{\mat{\Om}} \bar{\mA} \bar{\vx} }_1 \quad \mbox{s.t.} \\ 
 &\phantom{=} \quad \norm{\vy - \Rm \bar{\mPhi} \bar{\vx} }_2 \leq \tau,\quad \Rn\bar{\vx} \in \Rr \qandq \Rnc \bar{\vx} = 0. \nonumber
\end{align}
Let $\iota_{\norm{\vy - \cdot}\leq \tau}$, $\iota_{\Rr}$ and $\iota_{\{\zero\}}$ be the indicators of the sets $\enscond{\vz}{\norm{\vy - \vz} \leq \tau}$, $\Rr$ and $\{\zero\}$.
In order to match the required form, \eqref{eq:admm2_objective}, the problem \eqref{eq:l1analysis_ext} is then split in $J=3$ functions,
\eqn{
g_1(\bar{\vz}) &\eqdef \iota_{\norm{\vy - \cdot} \leq \tau} (\Rm \bar{\vz})  &\mbox{with }  &\mat{H}_1 \eqdef \bar{\mPhi}\\
g_2(\bar{\val}) &\eqdef \norm{ \bar{\mat{\Om}} \bar{\val}}_1  &\mbox{with }  &\mat{H}_2 \eqdef \bar{\mA} \\
g_3(\bar{\vx}) &\eqdef \iota_{\Rr}(\Rn \bar{\vx})+\iota_{\{\zero\}} (\Rnc \bar{\vx})  &\mbox{with }  &\mat{H}_3 \eqdef \Id.
}
The corresponding proximal operators all admit a very simple closed form expression that can be efficiently evaluated (see, \eg \cite{Combettes2009,Parikh2013} and references therein).
Moreover, we have
\begin{equation}
\label{eq:sum_of_operators}
\textstyle\sum_{j=1}^3 \mu_j {\mat{H}_j}^*\mat{H}_j = \mu_1\pa{ \bar{\mPhi}^*\bar{\mPhi} + \tfrac{\mu_2 + \mu_3}{\mu_1} \Id},
\end{equation}
which, as we will see, is easily invertible for both architectures.

For initializing the algorithm, we use a 3-D linear interpolation of $\mat{Y}$ in the 3-D $(u,v,\lambda)$ space to get an estimate $ \ten{Y}_{\rm lin} \in \RR^{m_u \times m_v \times n_{\lambda}}$. Let $\vy_{\rm lin} = \mathrm{vec}( \ten{Y}_{\rm lin})$ so that $\vy = \Rm \vy_{\rm lin}$ (but $ \vy_{\rm lin} \neq \Rm^* \vy$). We then use Tikhonov regularization,
\begin{equation}
\label{eq:tikhonov_x0}
\vx_{\rm init}  \eqdef \pa{ (\bar{\mPhi}\Rn)^*\bar{\mPhi}\Rn + \tau^2 \Id}^{-1} (\bar{\mPhi}\Rn)^* \vy_{\rm lin},
\end{equation}
that we practically solve using the conjugate gradients algorithm, \ie without matrix inversion.

%% file: inpaint.tex
\label{sec:msvi}

The first architecture, coined \acf{MSVI}, is presented in this section. We describe the formation and recording of measurements on the snapshot \ac{FP} sensor and the corresponding sensing matrix implementation. The description below is aligned with Fig.~\ref{fig:msvi_forward}.

\subsection{Image Formation Model}
\label{sec:msvi-sensing}

\begin{figure}[t]
\centering
\includegraphics[width=0.7\textwidth]{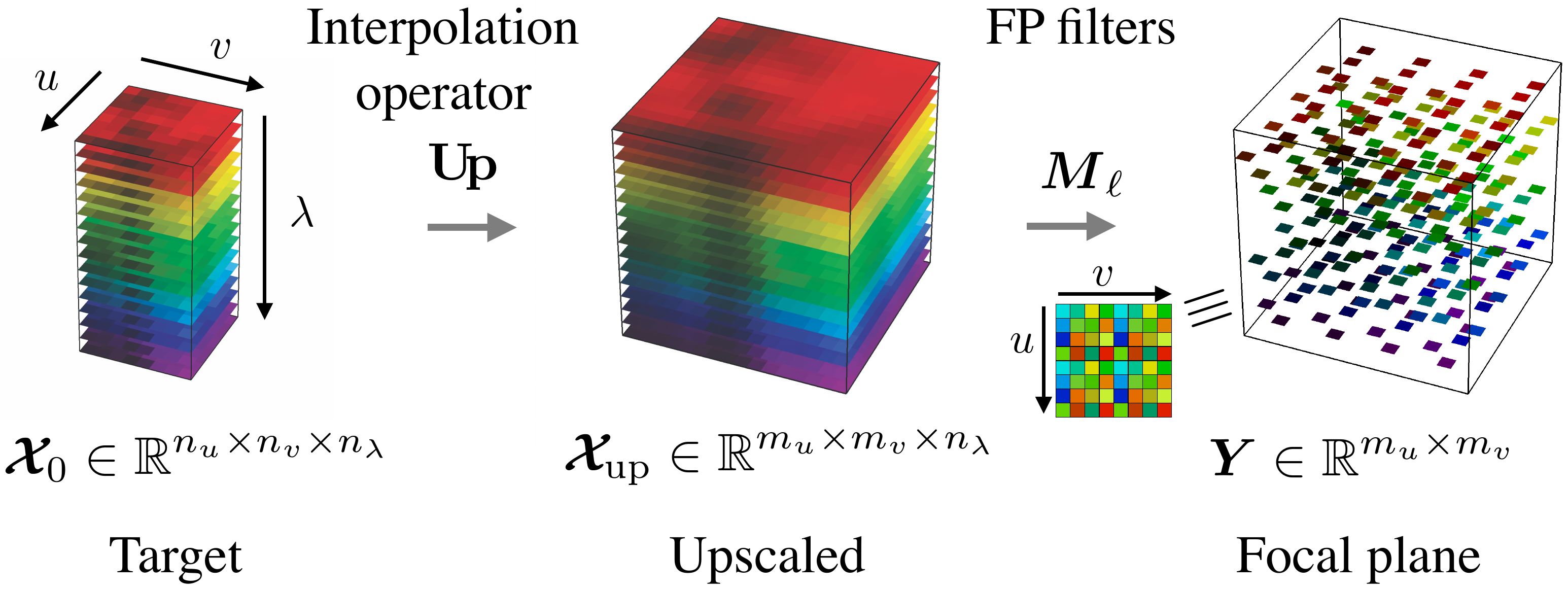}
\caption{\label{fig:msvi_forward} \ac{MSVI} forward model.}
\end{figure}

Our scheme allows us to choose, as a free parameter, the target spatial resolution, $n_u \times n_v$, of the target \ac{MS} volume, $\ten{X}_{\mkern-3mu 0} \in \mathbb{R}^{n_u \times n_v \times n_\lambda}$ (see Fig.~\ref{fig:msvi_forward}).
We choose to target a smaller resolution than the sensor resolution, $m_u \times m_v$, \ie $n_u\leq m_u$ and $n_v \leq m_v$, even though $m_um_v = m \leq n=n_un_vn_\la$. 
We assume that a spatial low-pass filter at appropriate frequency has been placed before the device so that the chosen resolution, $n_u \times n_v$, achieves the Nyquist rate of the resulting low-pass scene, $X_0(u,v,\la)$.
This practice is common for stabilizing demosaicking~\cite{Keelan2004}, \eg using birefringent filters~\cite{Acharya1994} or by slightly defocusing the objective lens.

 Let ${\ten{X}_{\mkern-3mu \rm up}} \in \RR^{m_u \times m_v \times n_\lambda}$ be an \emph{upscaled} version of the scene $\ten{X}_{\mkern-3mu 0}$, \ie matching the FPA pixel count $m_u \times m_v$. We can obtain this $\ten{X}_{\mkern-3mu \rm up}$ by using a smooth and separable interpolation function, \eg Lanczos, represented here by the linear operator, $\Up \in \RR^{m \times n_u n_v}$, applied to each band. Since $n_u \times n_v$ achieves the Nyquist rate, both $\ten{X}_{\mkern-3mu \rm up}$ and $\ten{X}_{\mkern-3mu 0}$ are lossless representations of $X_0(u,v,\lambda)$. 
This de-couples the number of FPA pixels, $m_u m_v$, from the target scene resolution, $n_u n_v$; we may choose the subsampling rate $m/n$ by changing one or the other.
In order to model the sensing operation and its relation with the upsampling $\Up$, we introduce the diagonal mask operator, $\mat{M}_\ell \in \{0,1\}^{m \times m}$, masking all \ac{FPA} pixels but the ones corresponding to the \ac{FP} filters of index ${\ell \in [n_\la]}$. 
Since every \ac{FPA} pixel is sampling exactly one band, we have $\sum_{\ell \in [n_\la]} \mat{M}_\ell = \Id$ and $\mat{M}_\ell \mat{M}_{\ell'} = \mat{0}$ for $\ell \neq \ell'$ so that the concatenation, $\mat{M} \eqdef (\mat{M}_1, \dots, \mat{M}_{n_\la}) \in \{0,1\}^{m \times n_\la m}$ is a \emph{restriction operator} in $\RR^{n_\la m}$ such that $\mat{M} \mat{M} ^* = \Id$.
The sensing matrix in \eqref{eq:forward_model_hsi} finally reads $\mPhi \eqdef \big( \mPhi_{1},\,\cdots,\mPhi_{n_{\lambda}} \big)$ with $\mPhi_{\ell} \eqdef \mat{M}_\ell \Up$. This forward model is schematized on Fig.~\ref{fig:msvi_forward}.

Set aside the clear affiliation with the inpainting problem in computer vision~\cite{Bertalmio2000,Ballester2001,Elad2005}, we can make links with the random basis ensembles (see, \eg \cite[Chapter 12]{Foucart2013} and references therein) in \ac{CS}. In the spectral direction, the sparsity basis is the DCT, which is \emph{maximally incoherent} with the canonical basis and therefore an optimal choice. On the other hand, in the spatial dimension, loosely speaking and ignoring upsampling, the sparsity ``basis'' is a wavelet transform which is \emph{not} maximally incoherent with the canonical basis. This intuitively justifies the study of the second method in Section~\ref{sec:msrc}. Rigorously extending the analogy to redundant wavelet analysis with the upsampling would require further work.

\subsection{Sensing matrix implementation}
\label{sec:msvi-matrix}
As explained in Section~\ref{sec:hsics-alg}, we can ease the computations by adding rows and columns to $\mPhi$ (see \eqref{eq:extended_Phi}). One natural choice is $\Rn = \Id$ and $\Rm = \mat{M}$, so that
$
\bar{\mPhi} = \bdiag_{n_\la} (\Up).
$
Therefore, $\bar{\mPhi}^*\bar{\mPhi} + \mu \Id$ is a separable sparse matrix which is easily pre-computed and fast to invert, for example with the conjugate gradients algorithm. Even though the gain is less obvious than in the \ac{MSRC} case discussed in Section~\ref{sec:msrc-matrix}, we found, empirically, that using this trick speeds up \ac{ADMM}'s convergence compared to the direct use of $\mPhi$. Let it be noted that in the case $m_u = n_u$ and $m_v = n_v$, \ie a subsampling rate of $1/n_\lambda$, we have $\Up = \Id$, which makes the inversion step as trivial as a scalar multiplication by $(1+\mu)^{-1}$.

%% file: msrconv.tex
\label{sec:msrc}

This section describes the \acf{MSRC} device. First, we discuss the image formation model, then some implementation aspects linked to important non-idealities, such as attenuation and diffraction. Finally, we discuss the numerical implementation of the sensing matrix, to be used in the recovery method of Section~\ref{sec:hsics-alg}.

\subsection{Image Formation Model}
\label{sec:msrc-sensing}

\begin{figure}
\centering
\includegraphics[width=0.65\textwidth]{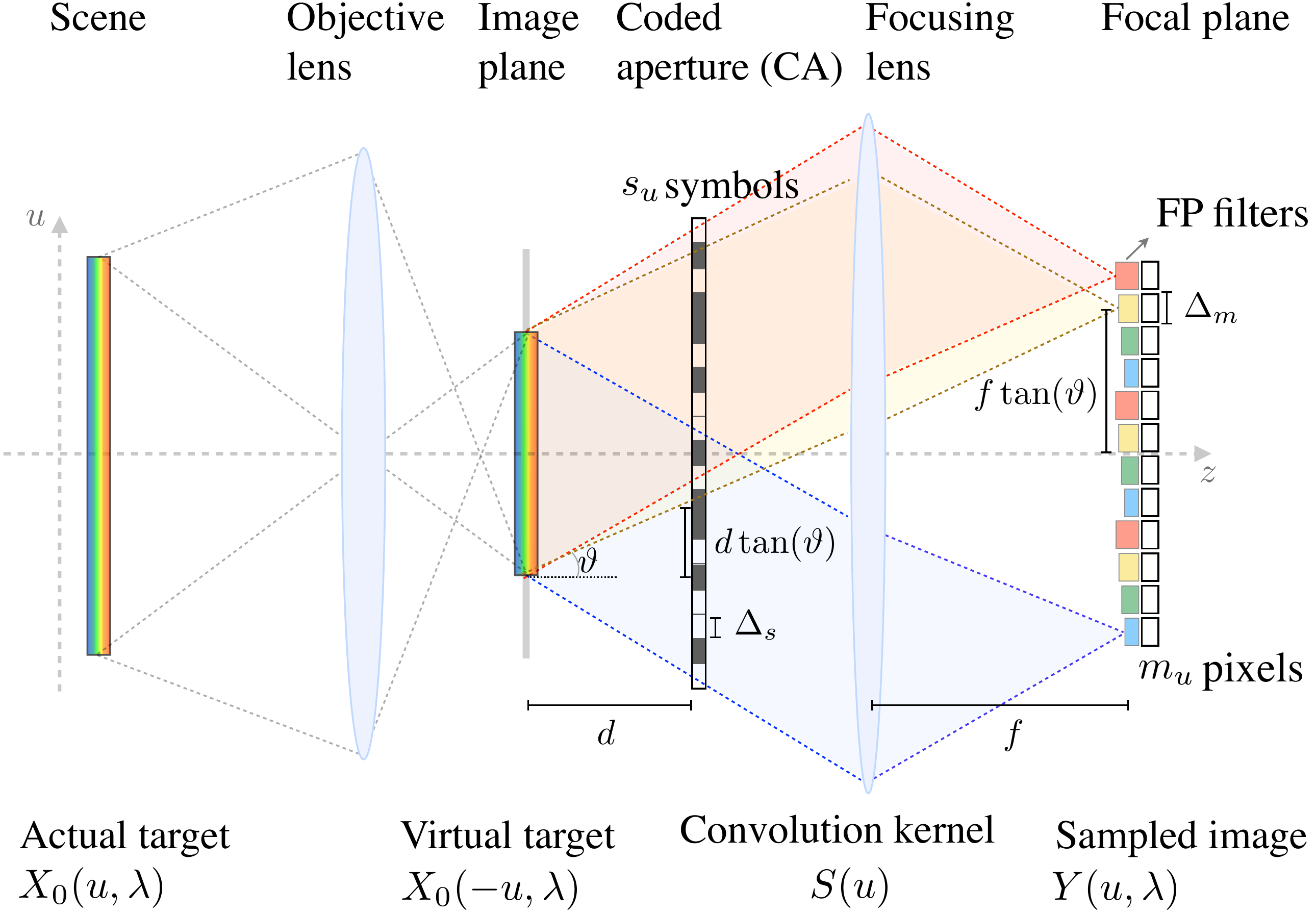}
\caption{\label{fig:msrc_forward} Geometry of the  \ac{MSRC}  optical path.}
\end{figure}

	We give here a description of the $\ac{MSRC}$ device, based on geometrical optics, depicted on Fig.~\ref{fig:msrc_forward}. This follows the ideas originally introduced by \cite{Bjorklund2013}. The difference, here, is that we use the \ac{FP} filtered sensor instead of a panchromatic sensor.

\subsubsection{Continuous model}	
	For a precise description, it is easier to use the continuous domain representation, $X_0(u,v,\lambda)$, of the object of interest.
	In order to lighten the notations, we will consider only one spatial dimension and use a simplified $X_0(u,\lambda)$ instead of $X_0(u,v,\lambda)$. Since everything is separable, the two dimensional extension is straightforward.
	 from which we consider that a virtual flipped source $X_0(-u,\lambda)$ radiates in all directions allowed by the aperture of the objective lens. It illuminates a random \ac{CA} with $s_u$ elements, at a distance $d$ along the optical axis. 
The \ac{CA}, with {\em aperture pitch} $\Delta_s$, is modeled by its transmittance function\footnote{In order to lighten the notations, we introduce the two following sampling functions, for any grid length, $n \in \NN$, and sampling rate (or pitch), $\De > 0$,
\eqn{
	 \de_i^n(\tfrac{u}{\De}) &\eqdef \delta \big( \tfrac{u}{\Delta} - (i-\tfrac{(n+1)}{2}) \big) \\
	 \rect_i^n(\tfrac{u}{\De}) &\eqdef \rect \big( \tfrac{u}{\Delta} - (i-\tfrac{n}{2}) \big)
}
where $\de(u)$ is Dirac's delta function and $\rect(u)$ is the boxcar function, \ie $1$ if $u\in [0,1)$ and $0$ elsewhere. They are defined so that for sampling indices $i \in [n]$, the sampling grid is always centered around $u=0$.},
\eql{
	S(u)  = \sum_{i=1}^{s_u} S_{i} \, \rect_i^{s_u}(\tfrac{u}{\De_s}). 
	\label{eq:discretized_CA}
}
	Its ${s_u}$ known {\em symbols} are $S_{i}$ with equal probability, modeling either transparent ($S_i = 1$) or opaque ($S_i = 0$) pixels.
	We assume that the \ac{CA} has negligible effect in the spectral domain. 
	The \ac{CA} is illuminated by replicas of the source, $X_0(u'(\vartheta)-u,\lambda)$ shifted by $u'(\vartheta) = d \tan (\vartheta)$ as bundles of parallel rays that propagate in the same directions, defined by the angle $\vartheta \in \left[-\nicefrac{\pi}{2},\nicefrac{\pi}{2}\right]$ w.r.t.  the optical axis.
	An ideal thin lens with focal length $f$, placed in front of the \ac{CA}, then focuses the modulated light on the sensor. All the rays with direction $\vartheta$ converge on the focal plane at $u''(\vartheta)= f \tan (\vartheta)$,
	\begin{align}
		Y(u''(\vartheta) ,\lambda) & = \int S(u)X_0(u'(\vartheta) - u,\lambda) \, \dif u \label{eq:measurements_continuous} \\ 
		& = [S \ast X_0](u'(\vartheta),\lambda) = [S \ast X_0](\tfrac{d}{f} u''(\vartheta),\lambda), \notag
	\end{align}
	with $\ast$ denoting, here, a continuous convolution.
	This defines the relationship with between $\Delta_s$, $f$, $d$ and the {\em pixel pitch} of the imaging sensor,
	$
	\Delta_m = \frac{\Delta_s f}{d}
	$. 
For $i \in [m_u]$, we choose to model the sampling function corresponding to the $i^{\mbox{\scriptsize th}}$ detector by $M_{i}(u,\lambda) = \delta^{m_u}_i ( \tfrac{u}{\Delta_m}) \, \delta(\lambda-\lambda_i)$. In this notation, we highlight the fact that the sensor is {\em spectrally-filtered}, \ie we assign a different  wavelength $\lambda_i$ depending on the pixel index $i$ (see Section~\ref{sec:fp_filters}). 
The $i^{\mbox{\scriptsize th}}$ measurement, is obtained as 
	\eql{
		y_{i} = \iint M_{i}(u,\lambda) [S \ast X_0](\tfrac{d}{f} u,\lambda) \, \dif u \, \dif \lambda,
		\label{eq:measurements_sampling}
	}
forming the discrete measurements vector, $\vy$. Note that this spectral filtering step is the continuous equivalent of the one described in Section~\ref{sec:msvi} and represented on the right part of Fig.~\ref{fig:msvi_forward}, but where $Y(u,\lambda)=[S \ast X_0](\tfrac{d}{f} u,\lambda)$ replaces $X_0(u,\lambda)$. There are $m_u$ sensor pixels, \ie $m_u$ shifts of the target on the \ac{CA}, which covers $n_u$ \ac{CA} elements. The \ac{CA} must therefore have $s_u = n_u+m_u-1$ elements to cover all recorded angles.

As explained in \cite{Bjorklund2013}, we can alter the \ac{CA} pattern and measurements vector so that the symbols of $S(u)$ effectively become $S_{i}\in\{-1,1\}$ instead of $S_{i}\in\{0,1\}$. We propose to either use two complementary patterns $S_+(u)$ and $S_-(u)$, where transparent pixels ($S_i = 1$) become opaque ($S_i = 0$) and vice versa, and subtract the corresponding measurements vectors,
		$\vy = \vy_+ - \vy_-$, 
or to subtract measurement made with a fully transparent aperture, $S_{\rm on}(u)$ (\ie $S_{i} = 1, \ \forall i$),  from $2\vy_+$, \ie $\vy = 2\vy_+ - \vy_{{\rm on}} $. This implies the use of a {\em programmable} \ac{CA} or a fixed mask that can easily be \emph{removed} for a full, non-coded acquisition (see Section~\ref{sec:msrc-nonidealities}). 
In the rest, we consider that the equivalent $S_{i}\in\{-1,1\}$ pattern is used.

\subsubsection{Discrete model}	
The discrete linear forward model of the optical processing chain stems from a particular discretization of the target volume. The most natural choice, in this instance, is to replace $X_0(u,\la)$, in \eqref{eq:measurements_sampling}, by its Dirac-sampled version,
\eql{
\label{eq:discrete_x0}
X_0^d(u,\la) \eqdef \sum_{i=1}^{n_u} \sum_{\ell=1}^{n_\la}  x_{0,i,\ell} 
\delta^{n_u}_i(\tfrac{u}{\De_s})
\delta(\la - \la_\ell),
}
where $x_{0,i,\ell}$ is a sample of the discrete target volume.
Note that the sampling functions of $M_{i}(u,\lambda)$ and $[S \ast X_0^d](\tfrac{d}{f}u,\la)$ are nicely aligned with each other, so that \eqref{eq:measurements_sampling} directly translates to the discrete model.
	Coming back to two discrete spatial dimensions, the discrete forward model thus reads
\eql{
	\label{eq:msrc_discrete_forward}
	\mat{Y} = \sum_{\ell = 1}^{n_\la} \mat{M}_\ell  ( {\mat{S}}\ \bar{\ast}\ \mat{X}_{0,\ell} ),
}
where $\mat{Y} \in \RR^{m_u \times m_v}$ is the array of recorded measurements; $\mat{M}_\ell(\cdot) : \RR^{m_u\times m_v} \rightarrow \RR^{m_u \times m_v}$ are the mask linear operators modeling the effect of the \ac{FP} filters (they correspond to the matrices $\mat{M}_\ell$ introduced in Section~\ref{sec:msvi}); the filter ${\mat{S}}\in \{-1,1\}^{s_u \times s_v}$ represents the discrete, 2-D version of $S(u)$; and $\mat{X}_{0,\ell} \in \RR^{n_u \times n_v}$ is the band of index $\ell$ of the full cube, $\ten{X}_{\mkern-3mu 0} \in \RR^{n_u\times n_v \times n_\la}$. 
The size, $s_u\times s_v$, of the \ac{CA} is chosen so that the \emph{valid} convolution (noted $\bar{\ast}$, see Section~\ref{sec:notations}) matches the size of the sensor, \ie $(s_u\!-\!n_u\!+\!1 , s_v\!-\!n_v\!+\!1) = (m_u , m_v)$.

\subsubsection{Multi-snapshot mode}
Since a total of $m_u \times m_v$ measurements is recorded by the sensor, the latter produces $\frac{m_u m_v}{n_\lambda}$ measurements per band. We consider the possibility of partitioning the acquisition of $\vy \in \RR^m$ by taking multiple snapshots, $\{\mat{Y}_{\!\!p}\}_{p \in [m_S] }$, with $m_S$ different aperture patterns, \ie $\{\mat{S}_{p}\}_{p \in [m_S] }$. Therefore, the total number of measurements becomes $m = m_u m_v m_S$. 
Taking multiple snapshots with different aperture patterns is expected to reduce the correlation between measurements. As the size of the FPA decreases, while keeping $m$ constant, the multi-snapshot device resembles more and more the single-pixel camera \cite{duarte:2008}, equivalent to setting $m_u=m_v=n_\lambda=1$.

\subsection{Non-idealities and practical considerations}
\label{sec:msrc-nonidealities}
The parallel compressive \ac{MSRC} scheme entails some additional concerns for its actual implementation. Hereafter, we explain the effect of diffraction and  a few other non-idealities. 

\subsubsection{Diffraction and Point Spread Function}

	As anticipated by \cite{Bjorklund2013}, the main optical-level limitation of this scheme is the impact of diffraction that occurs at the \ac{CA}. A single small square aperture, followed by a lens and illuminated by a plane wave, forms a diffraction pattern at the focal plane \cite[Chapter 4]{goodman2008introduction}. The effect of diffraction at the \ac{CA} is modeled as an optical filter whose \acf{PSF} is that pattern. 
The 2-D, wavelength-dependent, diffraction kernel has the following expression at the focal plane,
\eq{
H(u,v,\la) = a\, \sinc^2\pa{ u \frac{   \Delta_s }{ \la f } } \sinc^2\pa{ v \frac{   \Delta_s }{ \la f } },
}
where $a >0$ is an unimportant energy conservation constant (normalized afterwards).
This \ac{PSF} has a low-pass effect that limits the spatial bandwidth of the system, causing more correlation between measurements
and a decrease of performance. 
We again simplify the discussion to one spatial dimension.

Right before being sampled by the sensor, as in~\eqref{eq:measurements_sampling}, the ideal function $ Y(u,\la) $ is spatially convolved with $H(u,\la)$ as
\eql{
\label{eq:focal_plane_diffracted}
\tilde{Y}(u,\la) = [H \ast Y ](u,\la) = [\bar{H} \ast S \ast X_0](\tfrac{d}{f}u,\lambda),
}
with $\bar{H}(u,\la) \eqdef H( \tfrac{f}{d}u,\la) $, the kernel as equivalently viewed at the \ac{CA} scale. Note that this rescaled $\bar{H}(u,\la)$ does not physically appear at the \ac{CA} (since the diffraction pattern is only observed in the focal plane) but is only a notational trick allowing mathematical simplification. The expression for the discrete measurements \eqref{eq:measurements_sampling} becomes
\eql{
	\tilde{y}_{i} = \iint M_{i}(u,\lambda) [\bar{H} \ast S \ast X_0](\tfrac{d}{f}u,\lambda) \, \dif u \, \dif \lambda,
	\label{eq:measurements_sampling_diffracted}
}
In order to define the discrete sensing model, we inject \eqref{eq:discrete_x0} in \eqref{eq:measurements_sampling_diffracted} by replacing $X_0$ by $X_0^d$. After expanding the expression of $\tilde{y}_{i}$ (the details are omitted for brevity), one can verify that the result is completely equivalent to replacing the continuous kernel $\bar{H}(u)$ in \eqref{eq:measurements_sampling_diffracted} by a discretized version defined by
\eql{
\label{eq:diffraction_kernel_discretization}
\bar{H}^d(u) = \sum_{i=1}^{m_h} h_{i}(\la)  \, \de_i^{m_h}(\tfrac{u}{\De_s}). 
}
where the $m_h$ PSF samples $h_{i}(\la)$ are given by
\eql{
h_{i}(\la) \eqdef b 
\int
\bar{H}(u,\la)
\rect_i^{m_h} (\tfrac{u}{\Delta_s} )
\ \dif u.
\label{eq:diffraction_kernel_discrete}
}
The number, $m_h$, of kernel samples is determined by the size of the window in which they are significantly bigger than zero, and $b$ is a normalization factor such that $\sum_{i} h_{i}(\la) = 1$.  Note that sampling $\bar{H}(u,\la)$ with steps $\Delta_s$ is equivalent to sampling $H(u,\la)$ with steps $\Delta_m$. Also note that the sampling function, $\rect_i^{m_h} (\tfrac{u}{\Delta_s} )$, in \eqref{eq:diffraction_kernel_discrete} comes from the expression of $S(u)$ (see~\eqref{eq:discretized_CA}) but also depends on the chosen discretization $X_0^d$ and sampling functions~$M_i$, modeling the sensor pixels.

Let $\mat{H}_\ell \in \RR^{m_h\times m_h}$ be the 2-D discrete \ac{PSF}, by evaluating the 2-D generalization of \eqref{eq:diffraction_kernel_discrete} at wavelengths $\la_\ell$.
Using \eqref{eq:diffraction_kernel_discretization}, we can now adapt the 2-D multi-snapshot discrete model
as,
\eql{
	\label{eq:msrc_discrete_forward_diffracted}
	\mat{Y}_p = \sum_{\ell = 1}^{n_\la} \mat{M}_\ell  \big( (\mat{H}_\ell \ast \mat{S}_p) \ \bar{\ast}\ \mat{X}_{0,\ell} \big),
}
where we can pre-compute the diffracted, wavelength dependent aperture patterns, 
$\tilde{\mat{S}}_{p,\ell} \eqdef \mat{H}_\ell \ast \mat{S}_p$.
Since the size, $m_u\times m_v$, of the focal plane matches the valid convolution with a \ac{CA} of size $s_u \times s_v$, we can safely truncate the diffracted aperture pattern, $\tilde{\mat{S}}_{p,\ell}$, to an effective size of $s_u \times s_v$. 

Keep in mind that the modeled diffraction kernel is an approximation based on assumptions such as the use of a perfect thin lens, the fact that the object is an incoherent plane wave, etc. The actual \ac{PSF} of the system could be measured, for instance, using a pre-defined \ac{CA}
along with a point-like target light source to estimate the \ac{PSF} with a regularized inverse problem (see, \eg \cite{Gonzalez2016,Guerit2016}). 
Spatially-dependent \acp{PSF} could also be estimated with similar techniques. We leave this subject open to future investigation.

\subsubsection{Sizing example}
We can compute the size of the diffraction kernel as a function of the pixel pitches $\Delta_m$ and $\Delta_s$ and the focal length, $f$, which is constrained by the size of the lens and thus the size of the \ac{CA}.
Specifically, the diameter $D_{\rm lens}$ of the focusing lens must be bigger than the \ac{CA}, \ie
$D_{\rm lens} \geq \sqrt{2} \max\ens{s_u,s_v} \Delta_s$.
Moreover, practical lenses should have a sufficiently high F-number to avoid aberrations, \ie $\nicefrac{f}{D_{\rm lens} }\geq 0.5$.
We can characterize the width of the diffraction kernel on the sensor by the location of its first zeros, where the argument of the $\sinc^2(\cdot)$ is $1$, \ie in pixels,
\eq{
D_{\rm PSF,\la}  = 2\frac{\la f}{\Delta_m\Delta_s},
}
so that $H(\nicefrac{1}{2} D_{\rm PSF,\la} \Delta_m, \la)= \bar{H}(\nicefrac{1}{2} D_{\rm PSF,\la} \Delta_s, \la) =0$.
We thus apply this to the simulations parameters  of Section~\ref{sec:msrc-simu}, \ie $n_u = n_v = 256$, $m_u = m_v = 256$, so that $s_u = s_v = 511$.
Notice that the largest PSF width corresponds to the longest wavelength, $\la_{\rm max} = $ \unit{620}{nm}. Let $\Delta_s = $ \unit{80}{$\mu$m} so that the \ac{CA} is about \unit{41}{mm} wide and we must choose a lens of at least \unit{58}{mm} in diameter, with focal length $f \geq$ \unit{29}{mm}, \eg we can arbitrarily choose $f = $ \unit{40}{mm}. 
All these parameters being fixed, the PSF width on the focal plane (at $\la_{\rm max}\!\!=\,$\unit{620}{nm}) is \unit{620}{$\mu$m} and the number of pixels is determined by the pixel pitch $\Delta_m$ of the sensor, which also determines the distance~$d$.
\begin{figure}
\centering
\includegraphics[width=0.4\textwidth]{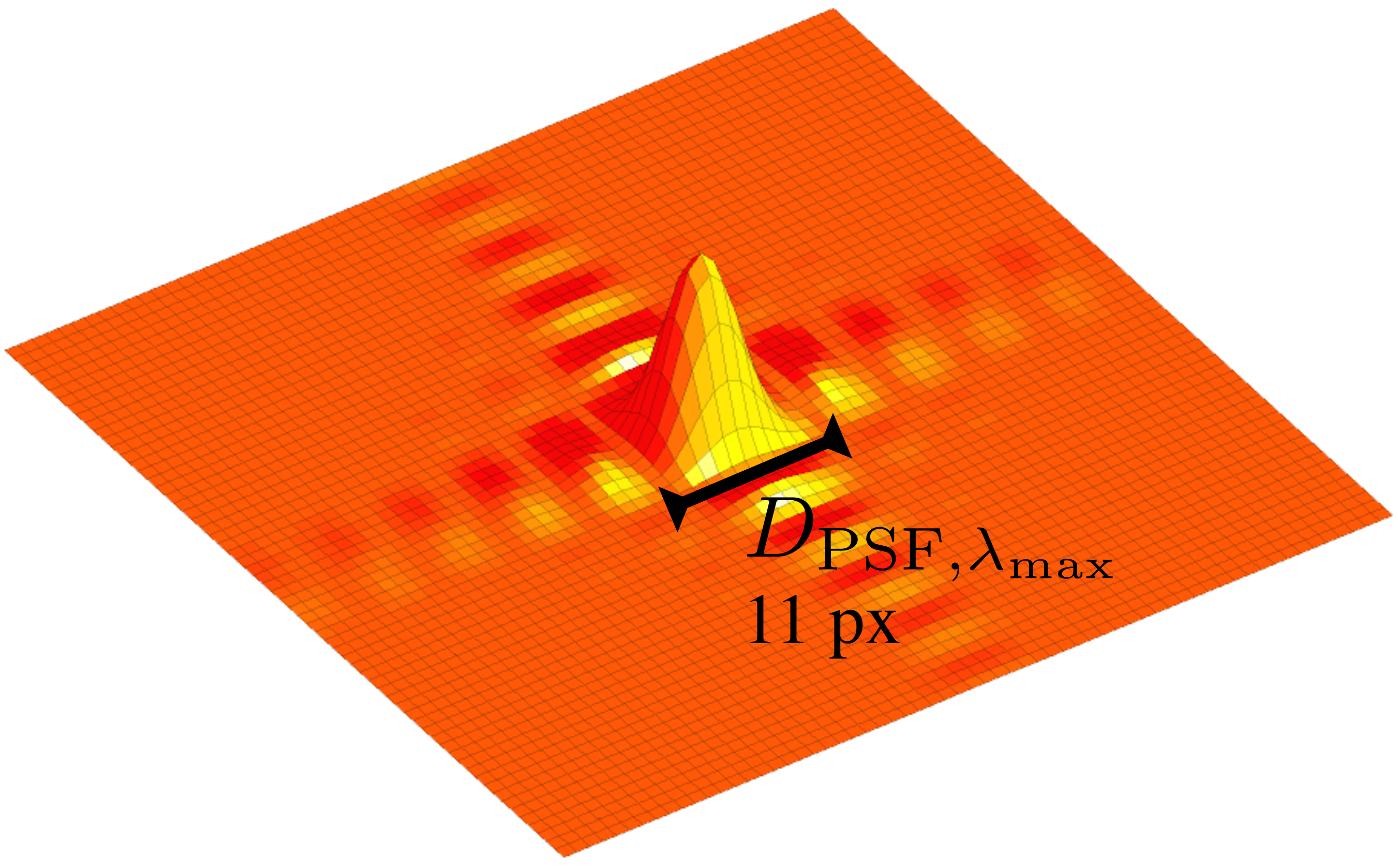}
\caption{\label{fig:psf_11px} \unit{11}{ pixels} wide diffraction \ac{PSF} ($\Delta_m = $ \unit{55}{$\mu m$}).}
\end{figure}
These parameters are incompatible with standard CMOS technology of $\Delta_m=$ \unit{5.5}{$\mu$m} used in \cite{geelen2014compact}. In that case, we get an impractical width of $D_{\rm PSF,\la_{\rm max}} = $ \unit{112}{ pixels}.
The workaround, proposed in \cite{Bjorklund2013}, of binning pixels together in macro-pixels is wasteful and defeats the purpose of the compressive architecture.
Another way of modifying the equivalent $\Delta_m$, requiring further investigation, is to magnify the sensor as viewed from the focusing lens.
For the simulations, intended as a proof of concept, we assume an effective magnification of 10 or 20 times the \unit{5.5}{$\mu$m} CMOS sensor. This leads to a width of respectively $D_{\rm PSF,\la_{\rm max}} = $ 11 and \unit{5}{ pixels} ($\Delta_m = $ 55 and \unit{110}{$\mu m$}). This \ac{PSF} is illustrated on Fig.~\ref{fig:psf_11px}.

\subsubsection{Other practical considerations}

We mention here a few other important challenges of the \ac{MSRC} design.
Beside the spectral differences mentioned in Section~\ref{sec:fp_filters}, manufacturing variability may introduce unknown gains in the sensor. 
Because each measurement provides information about the entire scene, this can highly limit the quality of the \ac{MSRC} reconstruction. 
In comparison, the effect of a corrupted measurement in the \ac{MSVI} design would be localized.
\emph{Blind} calibration techniques \cite{Cambareri2016,Ling2015} may help when \emph{direct} calibration is not possible.

Optical alignment is another important issue.
For instance, the distances $d$ and $f$ and the roll angle between the sensor and the \ac{CA} must be precisely set.
The choice of the lenses must also minimize chromatic and spherical aberrations.

Narrowband filtering considerably decreases the system's light throughput. 
Therefore, the implementation must limit further light attenuation. 
For instance, several possible choices exist for the \ac{CA}. A manufactured mask with physical holes provides the best light throughput but is not programmable. Pixels of a semi-transparent LCD \ac{SLM} have imperfect opacity or transparency. The same goes with reflective \emph{Liquid Crystal on Silicon (LCoS)} devices \cite{nagahara2010programmable}, paired with a polarizing beam splitter \cite{fowles2012introduction} that further dims the light. Despite their excellent light transmittance, \emph{Digital Micro-mirror Devices (DMD)}, such as the one used in the single-pixel camera \cite{duarte:2008}, are not suitable for being used out of focus, as uncertainty in the deflection angles (see, \eg \cite{dlp4500}) would translate into systematic error..

\subsection{Sensing matrix implementation}
\label{sec:msrc-matrix}

Based on the discrete model~\eqref{eq:msrc_discrete_forward_diffracted}, we can write the sensing matrix corresponding to the vectorized forward model~\eqref{eq:forward_model_hsi}. Let $\boldsymbol{\mathsf{S}}_{p,\ell} \in \RR^{m_um_v \times n_un_v}$ be the partial block circulant matrix which defines the valid convolution operator with $\tilde{\mat{S}}_{p,\ell}$, and let $\mat{M}_\ell \in \RR^{m_um_v \times m_um_v}$ be 
the matrix equivalent to $\mat{M}_\ell(\cdot)$ (\ie the same as in Section~\ref{sec:msvi-sensing}). First, notice that every band and every snapshot can  be processed separately by the submatrices
$\mPhi_{p,\ell} \eqdef \mat{M}_\ell \boldsymbol{\mathsf{S}}_{p,\ell}$ such that the vectorized form of~\eqref{eq:msrc_discrete_forward_diffracted} is $\vy_p = \sum_{\ell = 1}^{n_\la} \mPhi_{p,\ell} \vx_{0,\ell}$.
The sensing matrix, is thus the block matrix,
$
\mPhi = \left(\mPhi_{p,\ell} \right)_{p \in [m_S],\ell \in [n_\la]}$.
This follows from the natural order in which the $\vy_p $ and $\vx_{0,\ell}$ elements are stacked in the vectorized $\vy \in \RR^m$ and $\xz \in \RR^n$. Note that each $\mPhi_{p,\ell}$ is a masked (some rows are zeroed by  $\mat{M}_\ell$) random convolution which enjoys good \ac{CS} properties as explained in Section~\ref{sec:hsics_related_work}.
Let ${\bf R}_{m_um_v} \in \RR^{m_u m_v \times s_u s_v}$ be the restriction operator that selects the valid part, of size $m_u \times m_v$, of a \emph{circular} convolution of size $s_u \times s_v$. Similarly, let ${\bf R}_{n_un_v}^* \in \RR^{s_us_v \times n_un_v}$ be the zero-padding operator (adjoint of the restriction) whose output matches the size $s_u \times s_v$ of the circular convolution. Let $\mF \in \CC^{s_us_v \times s_us_v}$ be the 2-D \ac{DFT} and let $\mat{\Sigma}_{p,\ell} = \diag( \mF \vvec( \tilde{\mat{S}}_{p,\ell} ))$, \ie the diagonal matrix of the \ac{DFT} of $\tilde{\mat{S}}_{p,\ell}$. With all these ingredients, we can factorize,
\eq{
\mPhi_{p,\ell} =  \mat{M}_\ell {\bf R}_{m_um_v} \mF^* \mat{\Sigma}_{p,\ell} \mF {\bf R}_{n_un_v}^*.
}
The factors composing the full matrix $\mPhi$ thus read
$\Rn^* = \bdiag_{n_\la}({\bf R}_{n_un_v}^*)$, $\Rm = \bdiag_{m_S}(\tilde{\bf R}_{m_um_v})$,
with
\eq{
\tilde{\bf R}_{m_um_v} = \pa{\mat{M}_1 {\bf R}_{m_um_v}\, \dots, \mat{M}_{n_\la} {\bf R}_{m_um_v} },
}
and, denoting $\mF_{n_\la} = \bdiag_{n_\la}(\mF)$, $\mF_{m_Sn_\la} = \bdiag_{m_Sn_\la}(\mF)$ and the diagonal matrices $\tilde{\mat{\Sigma}}_p =  \bdiag(\mat{\Sigma}_{p,1}, \dots, \mat{\Sigma}_{p,n_\la} )$,
\eq{
\bar{\mPhi}= \mF_{m_Sn_\la}
 \begin{pmatrix} 
\tilde{\mat{\Sigma}}_1\\
\vdots	\\
 \tilde{\mat{\Sigma}}_{m_S}
 \end{pmatrix}  
\mF_{n_\la}.
}
With this factorization, $\bar{\mPhi}^*\bar{\mPhi} + \mu \Id$ is easily invertible. Indeed,  since $\mF_{m_Sn_\la} $ and $\mF_{n_\la}$ are unitary and $\tilde{\mat{\Sigma}}_p$ is diagonal, noting $\bar{\mat{\Sigma}}^2 = \sum_{p=1}^{m_S} \tilde{\mat{\Sigma}}_p^2$, we have
$
\bar{\mPhi}^*\bar{\mPhi} + \mu \Id
=  \mF_{n_\la}^* \pa{  \bar{\mat{\Sigma}}^2  + \mu \Id  } \mF_{n_\la}$.
Therefore, inverting $\bar{\mPhi}^*\bar{\mPhi} + \mu \Id$ is just equivalent to inverting the diagonal matrix, $  \bar{\mat{\Sigma}}^2   + \mu \Id  $.
Note that computing $\mPhi \vx$ for some input vector $\vx \in \RR^n$ requires computing $n_\la$ \acp{DFT} and $m_Sn_\la$ inverse \acp{DFT} of size $s_u\times s_v$. Similarly, computing $\mPhi^* \vz$ for some input vector $\vz \in \RR^m$ requires  $m_Sn_\la$ \acp{DFT} and $n_\la$ inverse \acp{DFT}. Comparatively, computing the inverse of $\bar{\mPhi}^*\bar{\mPhi} + \mu \Id$ is cheaper since it only requires $n_\la$ \acp{DFT}  and $n_\la$ inverse \acp{DFT}.

%% file: comp_exp.tex
\label{sec:numerics}

In this section, we present numerical results with experimental data for the \ac{MSVI} and with simulated acquisition to compare both \ac{MSVI} and \ac{MSRC} strategies. Experiments were performed in \textsc{Matlab} with the code provided in supplementary material \footnote{This paper has supplementary downloadable material available at https://github.com/kevd42/hsics\_tci .}.
In all the recoveries, we used $\mu_1 = 50 \frac{\rho}{\Vert \bar{\mPhi} \Vert^2}$, $\mu_2 = \mu_3 = \rho$, with $\rho = 40$. The dynamic range is normalized to $x_{\rm min} = 0$ and $x_{\rm max} = 1$. The tolerance of $\mathrm{tol} = 5.10^{-5}$ for the relative $\ell_2$-distance between two iterations was always reached in less than $n_{\mathrm{iter}} = 2000$ iterations. The $\mu_j$ parameters were manually tuned in order to reach a reasonably fast convergence, but they do not critically affect the recovery quality. The $\frac{\rho}{\Vert \bar{\mPhi} \Vert^2}$ factor in $\mu_1$ is a heuristic normalization of $\bar{\mPhi}^*\bar{\mPhi}$ compared to $\Id$ in \eqref{eq:sum_of_operators}.
For the initialization, we use a tolerance of $10^{-3}$ and a maximum of ten iterations.

\subsection{\ac{MSVI} Experiment}
\label{sec:msvi-exp}

\begin{figure}[t]
\centering
\includegraphics[width=0.5\textwidth]{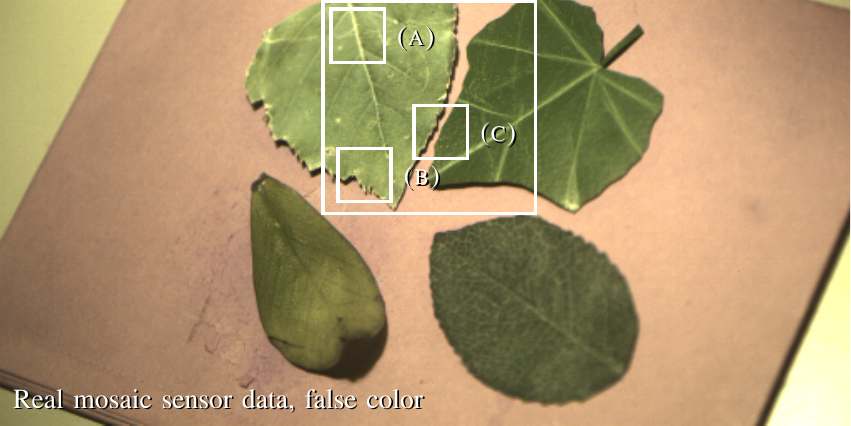}\\
\includegraphics[width=0.6\textwidth]{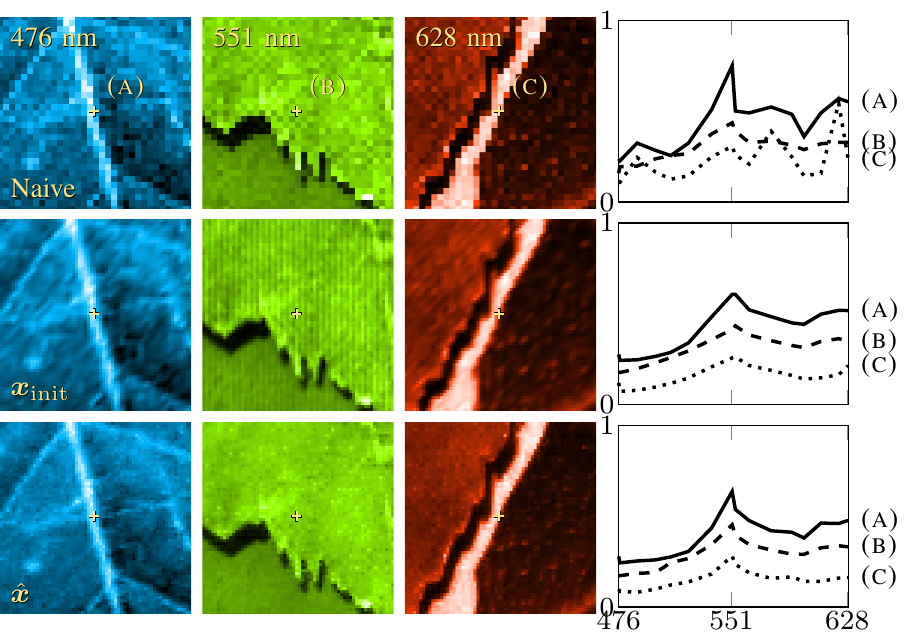}
\caption{\label{fig:experiment} Result of an experiments on real data acquired with imec's sensor. Qualitative comparison between naive demosaicking (nearest neighbor), the initialization \eqref{eq:tikhonov_x0}, and the proposed method. The top image is a false color nearest neighbor preview. The points  (\textsc{a}), (\textsc{b}) and (\textsc{c}) are pixels whose spectra are represented in the $4^{\mbox{\scriptsize th}}$ column.}
\end{figure}

On Fig.~\ref{fig:experiment}, we present the result based on experimental measurements of a test scene, observed with imec's mosaic sensor. 
This imager has a resolution of $1024\!\times\!2048$ pixels organized in a mosaic of $256\!\times\!512$ identical $4\!\times\!4$ macro-pixels; each with $n_\la = 16$ different \ac{FP} filters at wavelengths of visible light (as in Fig.~\ref{fig:sensor_layouts_mosaic}). For this experiment we restricted the measurements to a $512\!\times\!512$ region, depicted by the bigger white square on the false color image. The subsampling rate $m/n$, here is $1/4$, \ie we recover a volume with $256\!\times\!256\!\times\!16$ voxels. 
For setting $\tau$, we target a minimum measurements to residual ratio of $20\log_{10}(\norm{y}/\tau) = 40$dB.
The naive, super-pixel based, demosaicking method (top row), used in~\cite{geelen2014compact}, is clearly the worst. The middle row shows the result of the linear interpolation and Tikhonov regularization initialization method~\eqref{eq:tikhonov_x0}. Though we observe a clear improvement, a grid artifact,  already observed in \cite{Degraux2015a}, appears and is particularly visible on the 551 nm band. This artifact is removed with the proposed method (bottom row).  Without the exact filter calibration profiles and the ground truth spectra, we cannot, here, evaluate the spectral accuracy.

\subsection{Comparison of \ac{MSVI} and \ac{MSRC} on Synthetic Simulations}

\begin{figure*}[t]
\centering
\includegraphics[width=0.23\textwidth]{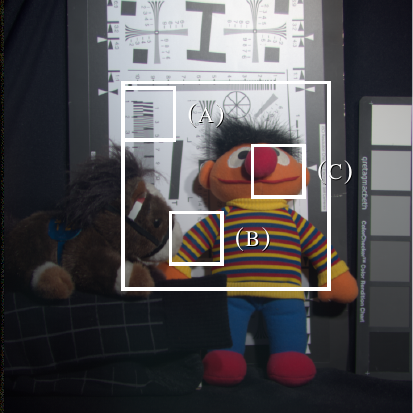}
\includegraphics[width=0.37\textwidth]{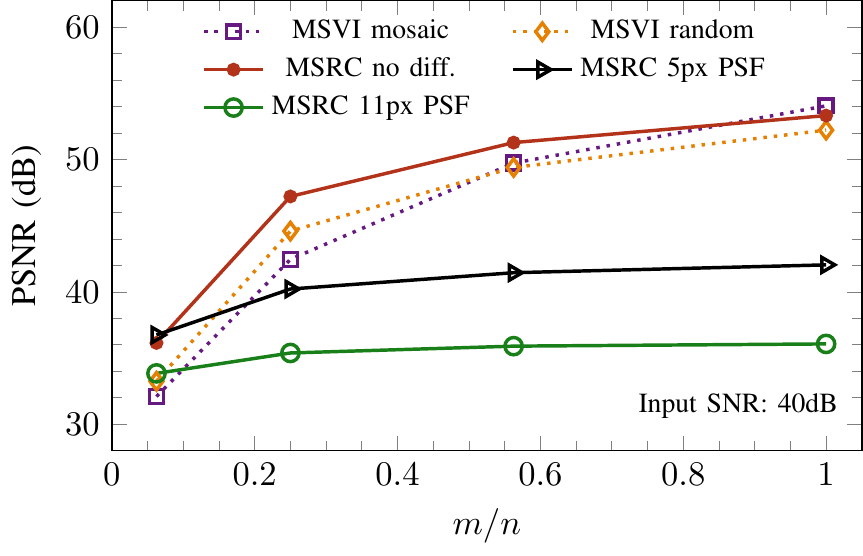}
\includegraphics[width=0.37\textwidth]{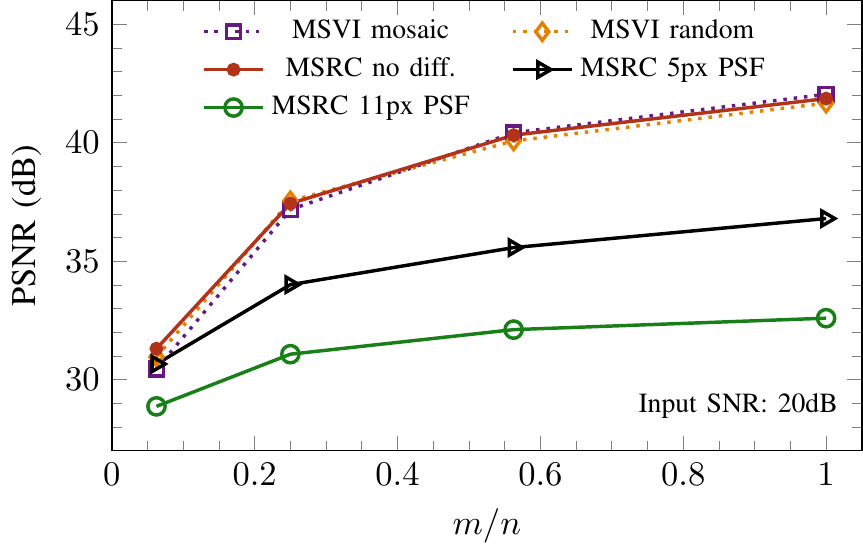}
\caption{\label{fig:quantitative_result} Results of the synthetic experiment. (Left) Ground truth in false color of the CAVE \cite{Yasuma2010} sample, \emph{chart and stuffed toy}. The bigger white square region indicates the part of the cube (\ac{ROI}) that was used in the experiment. The smaller white squares labelled (\textsc{a}), (\textsc{b}) and (\textsc{c}) correspond to zooms on features depicted on Fig.~\ref{fig:patches_results}. (Middle and right) Average PSNR over the whole dataset for two levels of input noise: 40dB (middle) and 20dB (right). The curves correspond to the \ac{MSVI} system (dotted lines) with mosaic (square purple) and  random (diamond orange) layouts, and to the \ac{MSRC} (plain lines) when diffraction is not modeled (red dots) and when $D_{\rm PSF,\la_{\rm max}}$ is 5 (black triangles) or 11 pixels (green circles). }
\end{figure*}

In the following, we use a \ac{MS} dataset to compare both strategies, quantitatively and qualitatively, on a series of controlled, synthetic simulations.
It comprises eight $256 \times 256 \times 16$ \ac{ROI} selected from the 32 multispectral $512 \times 512 \times 31$ volumes of the CAVE dataset \cite{Yasuma2010}. The spectral \ac{ROI} is 470 nm through 620 nm, matching \hyperlink{imec_sensor}{imec's sensor}. The spatial \ac{ROI} was manually chosen in each image to capture the most interesting features. The \emph{chart and stuffed toys} sample (\ac{ROI} centered at $(230,280)$), used for qualitative comparisons, is shown on Fig.~\ref{fig:quantitative_result} (left). The other samples, chosen to produce average PSNR curves, were  \emph{balloons} $(255,128)$, \emph{feathers} $(256,256)$, \emph{jelly beans} $(256,256)$, \emph{glass tiles} $(256,256)$, \emph{stuffed toys} $(256,256)$, \emph{superballs} $(200,236)$, and \emph{beads} $(256,256)$.
The middle and right plots on Fig.~\ref{fig:quantitative_result} show the average (over the eight dataset samples) reconstruction \acl{PSNR} ($\mathrm{PSNR} \eqdef -10 \log_{10}(\mathrm{MSE})$, where MSE stands for \acl{MSE})\acused{PSNR} in function of the subsampling rate $m/n$ for five different sensor configurations; two MSVI and three MSRC setups, each with two levels, 40 and 20dB (\ac{SNR}), of additive white gaussian noise on the measurements. The value of $\tau$ is determined by an oracle. Each simulation uses the corresponding sensing operator, $\mPhi$, for both generating the measurements and for reconstruction, \ie there is no model mismatch.
Fig.~\ref{fig:patches_results} shows the qualitative result of the chosen sample at $m/n = 1/16$, 40dB SNR. We chose this extreme low sampling rate under low input noise to expose the most obvious differences between sensing strategies and configurations.

\subsubsection{\ac{MSVI}}
\label{sec:msvi-simu}

Since $n$ is fixed by the dataset, we explore four \ac{MSVI} sensor sizes: $m_u = m_v \in \ens{256,512,768,1024}$. 
We test two different \ac{FP} filters configurations: the mosaic pattern (Fig.~\ref{fig:sensor_layouts_mosaic}) and the random pattern (Fig.~\ref{fig:sensor_layouts_random}).
At lower subsampling ratios, the random arrangement outperforms the mosaic sensor, particularly at high input SNR. This indicates that randomness mitigates aliasing caused by extreme subsampling. 
At Nyquist rate, mosaic beats random sampling, but both results are above 50dB and look visually perfect (not shown).

\begin{figure}[!ht]
\centering
\includegraphics[width=0.6\textwidth]{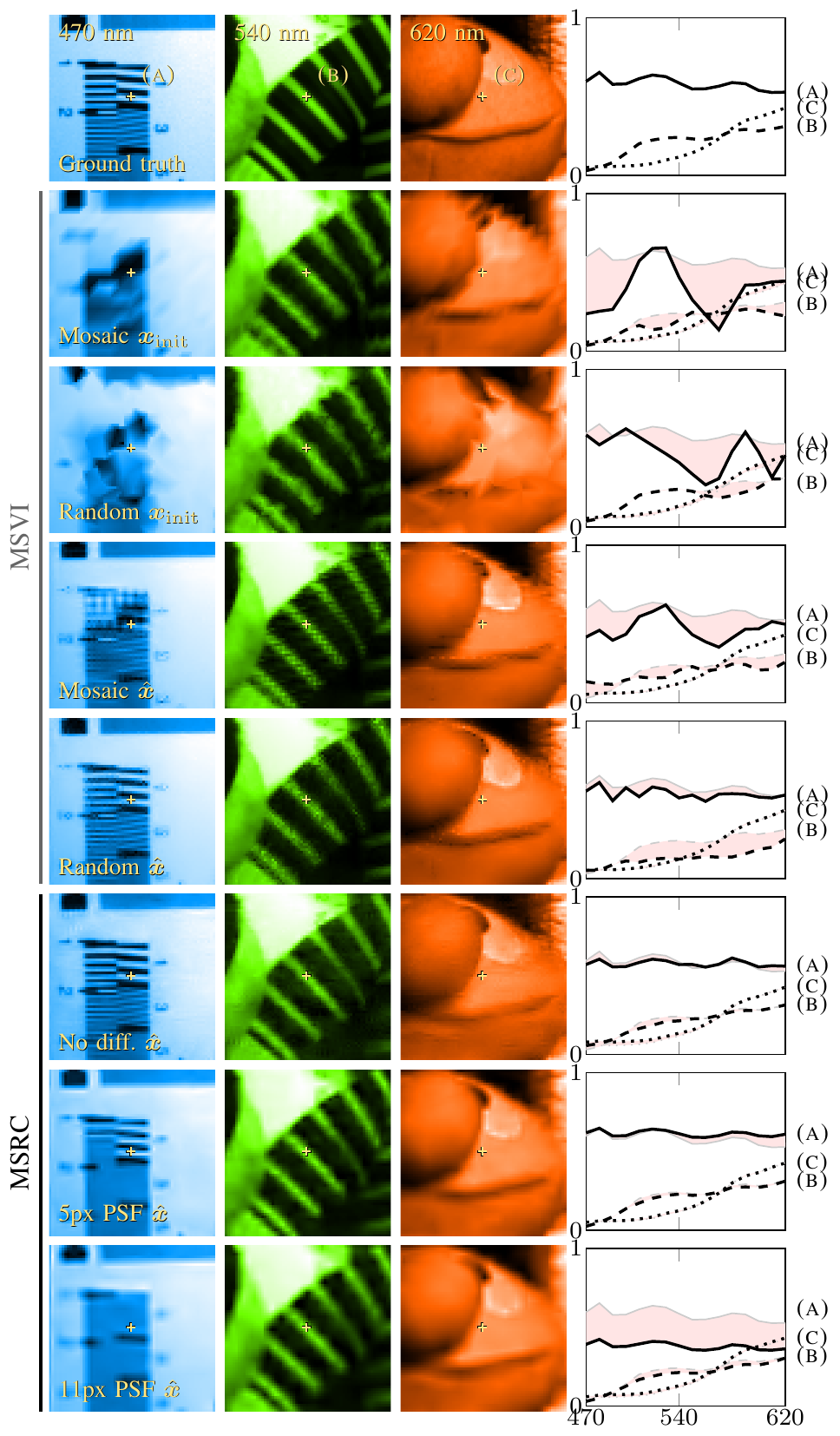}
\caption{\label{fig:patches_results} Qualitative results for the five sensor setups at $m/n = 1/16$, 40dB input SNR. The patches correspond to zooms on regions of the cube (Fig.~\ref{fig:quantitative_result}). The points  (\textsc{a}), (\textsc{b}) and (\textsc{c}) are pixels whose spectra are represented in the $4^{\mbox{\scriptsize th}}$ column. The light gray lines remind the ground truth and the red areas represent the error. The $2^{\mbox{\scriptsize nd}}$ and $3^{\mbox{\scriptsize rd}}$ rows show the results of linear interpolation of the mosaic and random MSVI measurements. The last five rows are the recovery results corresponding to each tested setup.}
\end{figure}

On Fig.~\ref{fig:patches_results}, we first show the initialization point as defined by \eqref{eq:tikhonov_x0}, \ie since $\bar{\mPhi} = \Id$, $\vx_{\rm init} = (1+\tau^2)^{-1} \vy_{\rm lin}$.  Despite being much faster than our iterative method, it gives visually scrambled results, particularly bad on the outer 470 nm and 620 nm bands where less data-points are available. The spectral error is particularly large for pixel (\textsc{A}) where the highly textured region destabilizes linear interpolation.
The results obtained by the proposed method, denoted $\xh$, are more accurate: edges and textures, \eg the horizontal bars of the chart and the stripes on the toy's sleeve are well resolved. The mosaic arrangement leads to a grid artifact as observed in Section~\ref{sec:msvi-exp}, whereas the random arrangement leads to seemingly unstructured artifacts and smaller spectral error areas, which concurs with the PSNR curves. 
In practice, a higher sampling rate, \eg $m/n = 1/4$, is preferable to mitigate those effects.

\subsubsection{\ac{MSRC}}
\label{sec:msrc-simu}

For testing the \ac{MSRC} strategy, the size of the \ac{FPA} is fixed to $m_u = m_v = 256$, so that $s_u = s_v = 511$.  To vary the sampling rate $m/n$, the number of snapshots is increased as $m_S \in \ens{1,4,9,16}$. Simulations with a unique snapshot and increasing \ac{FPA} size, omitted for the sake of conciseness, gave very close but slightly inferior results.
We compare the performances of a diffraction-free case with two cases where the diffraction kernel was respectively $D_{\rm PSF,\la_{\rm max}} = $ 5 and 11 pixels wide.
As expected, the global trend indicates that increasing the size of the \ac{PSF} decreases quality. Interestingly, at $m/n = 1/16$, the reconstruction PSNR of the diffraction-free case is on par with the 5 pixels case.

Fig.~\ref{fig:patches_results} suggests that the \ac{MSRC} method is suitable for extreme subsampling (compression). For example, the digits on the chart (first column) of the diffraction-free reconstruction are legible and artifacts are barely noticeable. The spectral error is also impressively small.
The performances rapidly decrease with diffraction and its spatial low-pass effect. As expected, the redundant wavelet prior is not able to recover the lost high-pass information. However, where the 11 pixels case gives pretty bad spectral accuracy, especially near edges, the 5 pixels case remains reasonably good at spectral reconstruction.

\subsubsection{Comparison}
In the ideal diffraction-free case under low noise, the \ac{MSRC} device provides a performance improvement of up to 4dB (for the $1/16$ subsampling rate) over the \ac{MSVI}. This justifies the present study on the feasibility of the \ac{MSRC} design.
However, at higher sampling rates, diffraction decreases quality, even with a 5 pixels PSF.
Note that the gap between \ac{MSVI}
and the ideal \ac{MSRC} falls to zero at 20dB of input SNR. 
For the qualitative comparison, we focus on the case, where \ac{MSRC} outperforms \ac{MSVI} on average, even with diffraction. 
\ac{MSRC} gives better spectral accuracy than \ac{MSVI} on the selected pixels, in particular pixels~(\textsc{a}) and~(\textsc{b}). Regarding spatial accuracy, noisy patterns appear between the stripes on the toy's sleeve with \ac{MSVI} reconstruction. However, the spatial high-frequency content, particularly visible on the chart patterns and digits, is affected by the diffraction kernel.

%% file: conclhsi.tex
\label{sec:conclusion}
Both strategies proposed in this paper use a \ac{MS} sensor with integrated \ac{FP} filters. Despite using the principles of \ac{CS}, they do \emph{not involve dispersive elements}.
Along with the conceptual optical design, for each device, we proposed an accurate forward model and a unified reconstruction procedure, formulated as a regularized inverse problem with an original spatio-spectral prior.
The particularity of \ac{MSRC}, compared to \ac{MSVI}, lies in the spatial mixing provided by an out-of-focus \ac{CA}, which allows higher compression ratios but, if not properly sized, entails adverse effects such as diffraction.

Through extensive numerical simulations, we explored different setups. We devised practical guidelines and highlighted limitations for both methods allowing to proceed towards an informed implementation.
In an ideally sized, low-noise, calibrated setup, \ac{MSRC} gives better performances with high compression. In other situations, factoring the cost of implementation and calibration, \ac{MSVI} should be preferred.

	\section*{Acknowledgments}
	The authors thank the Integrated Imagers group of imec (Leuven, Belgium) for supporting part of this design exploration during the second author's visit between February and August 2014.

%% file: hsics.bbl
\begin{thebibliography}{10}
\providecommand{\url}[1]{#1}
\csname url@samestyle\endcsname
\providecommand{\newblock}{\relax}
\providecommand{\bibinfo}[2]{#2}
\providecommand{\BIBentrySTDinterwordspacing}{\spaceskip=0pt\relax}
\providecommand{\BIBentryALTinterwordstretchfactor}{4}
\providecommand{\BIBentryALTinterwordspacing}{\spaceskip=\fontdimen2\font plus
\BIBentryALTinterwordstretchfactor\fontdimen3\font minus
  \fontdimen4\font\relax}
\providecommand{\BIBforeignlanguage}[2]{{%
\expandafter\ifx\csname l@#1\endcsname\relax
\typeout{** WARNING: IEEEtran.bst: No hyphenation pattern has been}%
\typeout{** loaded for the language `#1'. Using the pattern for}%
\typeout{** the default language instead.}%
\else
\language=\csname l@#1\endcsname
\fi
#2}}
\providecommand{\BIBdecl}{\relax}
\BIBdecl

\bibitem{Shippert2004}
P.~Shippert, ``{Why Use Hyperspectral Imagery?}'' \emph{Photogrammetric
  engineering and remote sensing}, vol.~70, pp. 377--379, 2004.

\bibitem{Tatzer2005}
P.~Tatzer, T.~Panner, M.~Wolf, and G.~Traxler, ``{Inline sorting with
  hyperspectral imaging in an industrial environment},'' N.~Kehtarnavaz and
  P.~A. Laplante, Eds., vol. 5671.\hskip 1em plus 0.5em minus 0.4em\relax
  International Society for Optics and Photonics, feb 2005, p. 162.

\bibitem{Hege2004}
E.~K. Hege, D.~O'Connell, W.~Johnson, S.~Basty, and E.~L. Dereniak,
  ``{Hyperspectral imaging for astronomy and space surviellance},'' in
  \emph{Optical Science and Technology, SPIE's 48th Annual Meeting}, S.~S. Shen
  and P.~E. Lewis, Eds., vol. 5159.\hskip 1em plus 0.5em minus 0.4em\relax
  International Society for Optics and Photonics, jan 2004, p. 380.

\bibitem{Gowen2007}
A.~A. Gowen, C.~P. O'Donnell, P.~J. Cullen, G.~Downey, and J.~M. Frias,
  ``{Hyperspectral imaging - an emerging process analytical tool for food
  quality and safety control},'' \emph{Trends in Food Science {\&} Technology},
  vol.~18, pp. 590--598, 2007.

\bibitem{Lu2014}
G.~Lu and B.~Fei, ``{Medical hyperspectral imaging: a review},'' \emph{Journal
  of Biomedical Optics}, vol.~19, p. 010901, 2014.

\bibitem{Whiting2006}
M.~L. Whiting, S.~L. Ustin, P.~Zarco-Tejada, A.~Palacios-Orueta, and V.~C.
  Vanderbilt, ``{Hyperspectral mapping of crop and soils for precision
  agriculture},'' W.~Gao and S.~L. Ustin, Eds., vol. 6298.\hskip 1em plus 0.5em
  minus 0.4em\relax International Society for Optics and Photonics, aug 2006,
  p. 62980B.

\bibitem{SellarBoreman2005}
R.~G. Sellar and G.~D. Boreman, ``{C}omparison of relative signal-to-noise
  ratios of different classes of imaging spectrometer,'' \emph{{A}pplied
  optics}, vol.~44, pp. 1614--1624, 2005.

\bibitem{HagenKudenov2013}
N.~Hagen and M.~W. Kudenov, ``{R}eview of snapshot spectral imaging
  technologies,'' \emph{{O}ptical {E}ngineering}, vol.~52, pp.
  090\,901--090\,901, 2013.

\bibitem{lambrechts2014cmos}
A.~Lambrechts, P.~Gonzalez, B.~Geelen, P.~Soussan, K.~Tack, and M.~Jayapala,
  ``{A CMOS-compatible, integrated approach to hyper- and multispectral
  imaging},'' \emph{2014 IEEE International Electron Devices Meeting}, pp.
  10.5.1--10.5.4, 2014.

\bibitem{geelen2014compact}
B.~Geelen, N.~Tack, and A.~Lambrechts, ``{A compact snapshot multispectral
  imager with a monolithically integrated per-pixel filter mosaic},'' in
  \emph{SPIE MOEMS-MEMS}.\hskip 1em plus 0.5em minus 0.4em\relax International
  Society for Optics and Photonics, 2014.

\bibitem{Bayer1976}
B.~E. Bayer, ``{Color Imaging Array. U. S. patent 3971 065},'' p.~10, 1976.

\bibitem{Donoho2006}
D.~L. Donoho, ``{Compressed sensing},'' \emph{IEEE Transactions on Information
  Theory}, vol.~52, pp. 1289--1306, 2006.

\bibitem{CandesWakin2008}
E.~J. Cand{\`e}s and M.~B. Wakin, ``{A}n introduction to compressive
  sampling,'' \emph{{S}ignal {P}rocessing {M}agazine, {IEEE}}, vol.~25, pp.
  21--30, 2008.

\bibitem{WillettMarciaNichols2011}
R.~M. Willett, R.~F. Marcia, and J.~M. Nichols, ``{C}ompressed sensing for
  practical optical imaging systems: a tutorial,'' \emph{{O}ptical
  {E}ngineering}, vol.~50, 2011.

\bibitem{ArceBradyCarinEtAl2014}
G.~Arce, D.~Brady, L.~Carin, H.~Arguello, and D.~Kittle, ``{C}ompressive coded
  aperture spectral imaging: {A}n introduction,'' \emph{{IEEE} {S}ignal
  {P}rocessing {M}agazine}, vol.~31, pp. 105--115, 2014.

\bibitem{GehmJohnBradyEtAl2007}
M.~Gehm, R.~John, D.~Brady, R.~Willett, and T.~Schulz, ``{S}ingle-shot
  compressive spectral imaging with a dual-disperser architecture,''
  \emph{{O}ptics {E}xpress}, vol.~15, pp. 14\,013--14\,027, 2007.

\bibitem{GehmBrady2015}
M.~E. Gehm and D.~J. Brady, ``{C}ompressive sensing in the {EO}/{IR},''
  \emph{{A}ppl. {O}pt.}, vol.~54, pp. C14--C22, Mar. 2015.

\bibitem{Candes2008}
E.~J. Cand{\`{e}}s, ``{The restricted isometry property and its implications
  for compressed sensing},'' \emph{Comptes Rendus Mathematique}, vol. 346, pp.
  1--4, 2008.

\bibitem{Rauhut2010a}
H.~Rauhut, ``{Compressive Sensing and Structured Random Matrices},'' in
  \emph{Radon Series Comp. Appl. Math}, M.~Fornasier, Ed.\hskip 1em plus 0.5em
  minus 0.4em\relax DE GRUYTER, jan 2010, pp. 1--94.

\bibitem{Degraux2015a}
K.~Degraux, V.~Cambareri, L.~Jacques, B.~Geelen, C.~Blanch, and G.~Lafruit,
  ``{Generalized inpainting method for hyperspectral image acquisition},'' in
  \emph{2015 IEEE International Conference on Image Processing (ICIP)}, vol.
  2015-Decem.\hskip 1em plus 0.5em minus 0.4em\relax IEEE, sep 2015, pp.
  315--319.

\bibitem{Degraux:iTWIST14}
K.~Degraux, V.~Cambareri, B.~Geelen, L.~Jacques, G.~Lafruit, and G.~Setti,
  ``{Compressive Hyperspectral Imaging by Out-of-Focus Modulations and
  Fabry-Perot Spectral Filters},'' in \emph{Proceedings of the second
  "international Traveling Workshop on Interactions between Sparse models and
  Technology" (iTWIST'14)}, 2014, pp. 21--23.

\bibitem{Boyd2010}
S.~Boyd, ``{Distributed Optimization and Statistical Learning via the
  Alternating Direction Method of Multipliers},'' \emph{Foundations and
  Trends{\textregistered} in Machine Learning}, vol.~3, pp. 1--122, 2010.

\bibitem{Almeida2013}
M.~S.~C. Almeida and M.~A.~T. Figueiredo, ``{Deconvolving Images With Unknown
  Boundaries Using the Alternating Direction Method of Multipliers},''
  \emph{IEEE Transactions on Image Processing}, vol.~22, pp. 3074--3086, aug
  2013.

\bibitem{SunKelly2009}
T.~Sun and K.~Kelly, ``{C}ompressive {S}ensing {H}yperspectral {I}mager,'' in
  \emph{Frontiers in Optics 2009/Laser Science XXV/Fall 2009 OSA Optics \&
  Photonics Technical Digest}.\hskip 1em plus 0.5em minus 0.4em\relax Optical
  Society of America, 2009, p. CTuA5.

\bibitem{WagadarikarPitsianisSunEtAl2009}
A.~A. Wagadarikar, N.~P. Pitsianis, X.~Sun, and D.~J. Brady, ``{V}ideo rate
  spectral imaging using a coded aperture snapshot spectral imager,''
  \emph{{O}ptics {E}xpress}, vol.~17, pp. 6368--6388, 2009.

\bibitem{WagadarikarJohnWillettEtAl2008}
A.~Wagadarikar, R.~John, R.~Willett, and D.~Brady, ``{S}ingle disperser design
  for coded aperture snapshot spectral imaging,'' \emph{{A}pplied {O}ptics},
  vol.~47, pp. B44--B51, 2008.

\bibitem{WuMirzaArceEtAl2011}
Y.~Wu, I.~O. Mirza, G.~R. Arce, and D.~W. Prather, ``{D}evelopment of a
  digital-micromirror-device-based multishot snapshot spectral imaging
  system,'' \emph{{O}ptics {L}etters}, vol.~36, pp. 2692--2694, 2011.

\bibitem{CorreaArguelloArce2014}
C.~V. Correa, H.~Arguello, and G.~R. Arce, ``{C}ompressive spectral imaging
  with colored-patterned detectors,'' in \emph{Acoustics, Speech and Signal
  Processing (ICASSP), 2014 IEEE International Conference on}.\hskip 1em plus
  0.5em minus 0.4em\relax IEEE, 2014, pp. 7789--7793.

\bibitem{CorreaArguelloArce2015}
------, ``{S}napshot colored compressive spectral imager,'' \emph{{JOSA} {A}},
  vol.~32, pp. 1754--1763, 2015.

\bibitem{AugustVachmanRivensonEtAl2013}
Y.~August, C.~Vachman, Y.~Rivenson, and A.~Stern, ``{C}ompressive hyperspectral
  imaging by random separable projections in both the spatial and the spectral
  domains,'' \emph{{A}pplied optics}, vol.~52, pp. D46--D54, 2013.

\bibitem{Fowler2014}
J.~E. Fowler, ``{C}ompressive pushbroom and whiskbroom sensing for
  hyperspectral remote-sensing imaging,'' in \emph{Image Processing (ICIP),
  2014 IEEE International Conference on}.\hskip 1em plus 0.5em minus
  0.4em\relax IEEE, 2014, pp. 684--688.

\bibitem{GeelenBlanchGonzalezEtAl2015}
B.~Geelen, C.~Blanch, P.~Gonzalez, K.~Tack, and A.~Lambrechts, ``{A} tiny
  {VIS}-{NIR} snapshot multispectral camera,'' in \emph{SPIE OPTO}, no.
  937414.\hskip 1em plus 0.5em minus 0.4em\relax International Society for
  Optics and Photonics, 2015.

\bibitem{duarte:2008}
J.~K. Romberg, M.~F. Duarte, M.~A. Davenport, D.~Takhar, J.~N. Laska, T.~Sun,
  K.~F. Kelly, and R.~G. Baraniuk, ``{Single-Pixel Imaging via Compressive
  Sampling},'' \emph{Signal Processing Magazine, IEEE}, vol.~25, pp. 83--91,
  mar 2008.

\bibitem{Lustig2007}
M.~Lustig, D.~Donoho, and J.~M. Pauly, ``{Sparse MRI: The application of
  compressed sensing for rapid MR imaging},'' \emph{Magnetic Resonance in
  Medicine}, vol.~58, pp. 1182--1195, 2007.

\bibitem{willett2011compressed}
R.~M. Willet, R.~F. Marcia, and J.~M. Nichols, ``{Compressed sensing for
  practical optical imaging systems: a tutorial},'' \emph{Optical Engineering},
  vol.~50, p. 072601, jul 2011.

\bibitem{dicke1968scatter}
R.~H. Dicke, ``{Scatter-Hole Cameras for X-Rays and Gamma Rays},'' \emph{The
  Astrophysical Journal}, vol. 153, p. L101, aug 1968.

\bibitem{fenimore1978coded}
E.~E. Fenimore and T.~M. Cannon, ``{Coded aperture imaging with uniformly
  redundant arrays},'' \emph{Applied Optics}, vol.~17, p. 337, feb 1978.

\bibitem{Romberg2009}
J.~Romberg, ``{Compressive Sensing by Random Convolution},'' \emph{SIAM Journal
  on Imaging Sciences}, vol.~2, pp. 1098--1128, jan 2009.

\bibitem{Rauhut2012}
H.~Rauhut, J.~Romberg, and J.~A. Tropp, ``{Restricted isometries for partial
  random circulant matrices},'' \emph{Applied and Computational Harmonic
  Analysis}, vol.~32, pp. 242--254, mar 2012.

\bibitem{Jacques2009}
L.~Jacques, P.~Vandergheynst, A.~Bibet, V.~Majidzadeh, A.~Schmid, and
  Y.~Leblebici, ``{CMOS compressed imaging by Random Convolution},'' in
  \emph{2009 IEEE International Conference on Acoustics, Speech and Signal
  Processing}.\hskip 1em plus 0.5em minus 0.4em\relax IEEE, apr 2009, pp.
  1113--1116.

\bibitem{Marcia2009}
R.~F. Marcia, Z.~T. Harmany, and R.~M. Willett, ``{Compressive coded aperture
  imaging},'' in \emph{IS{\&}T/SPIE Electronic Imaging}, C.~A. Bouman, E.~L.
  Miller, and I.~Pollak, Eds., feb 2009.

\bibitem{Bjorklund2013}
T.~Bj{\"{o}}rklund and E.~Magli, ``{A parallel compressive imaging architecture
  for one-shot acquisition},'' in \emph{2013 Picture Coding Symposium, PCS 2013
  - Proceedings}, nov 2013, pp. 65--68.

\bibitem{chan2008single}
W.~L. Chan, K.~Charan, D.~Takhar, K.~F. Kelly, R.~G. Baraniuk, and D.~M.
  Mittleman, ``{A single-pixel terahertz imaging system based on compressed
  sensing},'' \emph{Applied Physics Letters}, vol.~93, p. 121105, sep 2008.

\bibitem{Cambareri2016}
V.~Cambareri and L.~Jacques, ``{A Greedy Blind Calibration Method for
  Compressed Sensing with Unknown Sensor Gains},'' \emph{ArXiv e-prints
  arXiv:1610.02851}, pp. 1--6, 2016.

\bibitem{Ling2015}
S.~Ling and T.~Strohmer, ``{Self-calibration and biconvex compressive
  sensing},'' \emph{Inverse Problems}, vol.~31, p. 115002, nov 2015.

\bibitem{Fabry1901}
C.~Fabry and A.~Perot, ``{On a New Form of Interferometer},'' \emph{The
  Astrophysical Journal}, vol.~13, p. 265, may 1901.

\bibitem{Hernandez1986}
G.~Hernandez, \emph{{Fabry–Perot Interferometers: Cambridge Studies in Modern
  Optics}}.\hskip 1em plus 0.5em minus 0.4em\relax Cambridge University Press,
  may 1986, vol.~3.

\bibitem{geelen2013snapshot}
B.~Geelen, N.~Tack, and A.~Lambrechts, ``{A snapshot multispectral imager with
  integrated tiled filters and optical duplication},'' in \emph{Spie
  Moems-Mems}, G.~von Freymann, W.~V. Schoenfeld, and R.~C. Rumpf, Eds., vol.
  8613, mar 2013, p. 861314.

\bibitem{Elad2006}
M.~Elad, P.~Milanfar, and R.~Rubinstein, ``{Analysis versus synthesis in signal
  priors},'' \emph{European Signal Processing Conference}, vol.~23, pp.
  947--968, 2006.

\bibitem{Candes2011c}
E.~J. Cand{\`{e}}s, Y.~C. Eldar, D.~Needell, and P.~Randall, ``{Compressed
  sensing with coherent and redundant dictionaries},'' \emph{Applied and
  Computational Harmonic Analysis}, vol.~31, pp. 59--73, 2011.

\bibitem{Nam2013}
S.~Nam, M.~Davies, M.~Elad, and R.~Gribonval, ``{The cosparse analysis model
  and algorithms},'' \emph{Applied and Computational Harmonic Analysis},
  vol.~34, pp. 30--56, jan 2013.

\bibitem{Mallat2009}
S.~Mallat, \emph{{A Wavelet Tour of Signal Processing}}.\hskip 1em plus 0.5em
  minus 0.4em\relax Elsevier, 2009.

\bibitem{Starck2007}
J.~L. Starck, J.~Fadili, and F.~Murtagh, ``{The undecimated wavelet
  decomposition and its reconstruction},'' \emph{IEEE Transactions on Image
  Processing}, vol.~16, pp. 297--309, 2007.

\bibitem{Glowinski1975}
R.~Glowinski and A.~Marroco, ``{Sur l'Approximation, par Elements d'Ordre un,
  et la Resolution, par Penalisation-Dualit{\'{e}}, d'une Classe de Problemes
  de {\{}D{\}}irichlet non Lineares},'' \emph{Revue Fran{\c{c}}aise
  d'Automatique, Informatique, et Recherche Operationelle}, vol. 9(R-2), pp.
  41--76, 1975.

\bibitem{Gabay1976}
D.~Gabay and B.~Mercier, ``{A dual algorithm for the solution of nonlinear
  variational problems via finite element approximation},'' \emph{Computers
  {\&} Mathematics with Applications}, vol.~2, pp. 17--40, 1976.

\bibitem{Combettes2009}
P.~L. Combettes and J.-C. Pesquet, ``{Proximal Splitting Methods in Signal
  Processing},'' in \emph{Fixed-Point Algorithms for Inverse Problems in
  Science and Engineering}, H.~H. Bauschke, R.~S. Burachik, P.~L. Combettes,
  V.~Elser, D.~R. Luke, and H.~Wolkowicz, Eds.\hskip 1em plus 0.5em minus
  0.4em\relax Springer New York, 2011, pp. 185--212.

\bibitem{Parikh2013}
N.~Parikh and S.~P. Boyd, ``{Proximal Algorithms},'' \emph{Foundations and
  Trends in Optimization}, vol.~1, pp. 123--231, 2013.

\bibitem{Keelan2004}
B.~W. Keelan, \emph{{Handbook of image quality : characterization and
  prediction}}.\hskip 1em plus 0.5em minus 0.4em\relax New York: Marcel Dekker,
  2002.

\bibitem{Acharya1994}
T.~Acharya, K.~Bhattacharya, and A.~Ghosh, ``{Optical processing using a
  birefringence-based spatial filter},'' \emph{Journal of Modern Optics},
  vol.~41, pp. 979--986, 1994.

\bibitem{Bertalmio2000}
M.~Bertalmio, G.~Sapiro, V.~Caselles, and C.~Ballester, ``{Image inpainting},''
  in \emph{Proceedings of the 27th annual conference on Computer graphics and
  interactive techniques - SIGGRAPH '00}.\hskip 1em plus 0.5em minus
  0.4em\relax New York, New York, USA: ACM Press, 2000, pp. 417--424.

\bibitem{Ballester2001}
C.~Ballester, M.~Bertalmio, V.~Caselles, G.~Sapiro, and J.~Verdera,
  ``{Filling-in by joint interpolation of vector fields and gray levels},''
  \emph{IEEE Transactions on Image Processing}, vol.~10, pp. 1200--1211, 2001.

\bibitem{Elad2005}
M.~Elad, J.-L. Starck, P.~Querre, and D.~L. Donoho, ``{Simultaneous cartoon and
  texture image inpainting using morphological component analysis (MCA)},''
  \emph{Applied and Computational Harmonic Analysis}, vol.~19, pp. 340--358,
  2005.

\bibitem{Foucart2013}
S.~Foucart and H.~Rauhut, ``{A mathematical introduction to compressive
  sensing},'' \emph{Appl. Numer. Harmon. Anal. Birkh{\"{a}}user, Boston}, 2013.

\bibitem{goodman2008introduction}
J.~W. Goodman, \emph{{Introduction to Fourier Optics}}.\hskip 1em plus 0.5em
  minus 0.4em\relax Roberts and Company Publishers, 2005.

\bibitem{Gonzalez2016}
A.~Gonz{\'{a}}lez, V.~Delouille, and L.~Jacques, ``{Non-parametric PSF
  estimation from celestial transit solar images using blind deconvolution},''
  \emph{Journal of Space Weather and Space Climate}, vol.~6, p.~A1, jan 2016.

\bibitem{Guerit2016}
S.~Gu{\'{e}}rit, A.~Gonz{\'{a}}lez, A.~Bol, J.~A. Lee, and L.~Jacques, ``{Blind
  Deconvolution of PET Images using Anatomical Priors},'' \emph{Proceedings of
  the third "international Traveling Workshop on Interactions between Sparse
  models and Technology" iTWIST'16}, aug 2016.

\bibitem{nagahara2010programmable}
H.~Nagahara, C.~Zhou, T.~Watanabe, H.~Ishiguro, and S.~K. Nayar,
  ``{Programmable aperture camera using LCoS},'' in \emph{Lecture Notes in
  Computer Science (including subseries Lecture Notes in Artificial
  Intelligence and Lecture Notes in Bioinformatics)}, vol. 6316 LNCS, 2010, pp.
  337--350.

\bibitem{fowles2012introduction}
G.~R. Fowles, ``{Introduction to Modern Optics},'' \emph{American Journal of
  Physics}, vol.~36, p. 770, 1968.

\bibitem{dlp4500}
{Texas Instrument}, ``Dlp4500 0.45 wxga dmd,'' Part datasheet, April 2013
  (revised January 2016).

\bibitem{Yasuma2010}
F.~Yasuma, T.~Mitsunaga, D.~Iso, and S.~K. Nayar, ``{Generalized assorted pixel
  camera: Postcapture control of resolution, dynamic range, and spectrum},''
  \emph{IEEE Transactions on Image Processing}, vol.~19, pp. 2241--2253, 2010.

\end{thebibliography}
